\documentclass[10pt,twocolumn,letterpaper]{article}

\usepackage{graphicx}
\usepackage{subcaption}
\usepackage{float}
\usepackage[justification=raggedright]{caption}	
\usepackage{lscape}                                         

\usepackage[lined,ruled,linesnumbered]{algorithm2e}

\usepackage{booktabs}                   
\usepackage{multirow}

\usepackage{paralist}
\usepackage{enumitem}

\usepackage{bm}                          
\usepackage{epsfig}                      
\usepackage{graphicx}                  
\usepackage{times}
\usepackage{mathptmx}
\usepackage{mathtools}
\usepackage{amssymb,amsmath}   

\usepackage{units}
\usepackage{color}

\usepackage{comment}

\usepackage{url}  
\usepackage[pagebackref=true,breaklinks=true,letterpaper=true,colorlinks,bookmarks=false]{hyperref}

\usepackage{xspace}
\usepackage[table]{xcolor}
\usepackage{setspace}

\def\etal{et~al.\_}			  

\DeclareMathOperator*{\argmax}{\arg\!\max}

\newcommand{\mpage}[2]
{
\begin{minipage}{#1\linewidth}\centering
#2
\end{minipage}
}

\newlength\paramargin
\newlength\figmargin
\newlength\secmargin
\newlength\figcapmargin

\newcommand{\secref}[1]{Section~\ref{sec:#1}}
\newcommand{\figref}[1]{Figure~\ref{fig:#1}} 
\newcommand{\tabref}[1]{Table~\ref{tab:#1}}

\newcommand{\Paragraph}[1]{\vspace{\paramargin}\paragraph{#1}}

\long\def\ignorethis#1{}
\newcommand {\pohan}[1]{{\color{blue}\textbf{Po-Han: }#1}\normalfont}
\newcommand {\jiabin}[1]{{\color{red}\textbf{Jia-Bin: }#1}\normalfont}
\newcommand {\johannes}[1]{{\color{magenta}\textbf{Johannes: }#1}\normalfont}
\newcommand {\kevin}[1]{{\color{cyan}\textbf{Kevin: }#1}\normalfont}
\newcommand {\ahuja}[1]{{\color{green}\textbf{Ahuja: }#1}\normalfont}

\newcommand{\final}{0}
\ifthenelse{\equal{\final}{1}}
{
\renewcommand{\pohan}[1]{}
\renewcommand{\jiabin}[1]{}
\renewcommand{\johannes}[1]{}
\renewcommand{\kevin}[1]{}
\renewcommand{\ahuja}[1]{}
}
{}

\def\xi{\mathbf{x}_i}

\graphicspath{{figure}, {example}}

\usepackage{cvpr}
\usepackage{subcaption}

\cvprfinalcopy 

\def\cvprPaperID{2703} 

\ifcvprfinal\fi

\begin{document}

\title{DeepMVS: Learning Multi-view Stereopsis}

\author{
Po-Han Huang\textsuperscript{1}\\
\and
Kevin Matzen\textsuperscript{2}\\
\and
Johannes Kopf\textsuperscript{2}\\
\and
Narendra Ahuja\textsuperscript{1}\\
\and 
Jia-Bin Huang\textsuperscript{3}\\
\and
\textsuperscript{1}University of Illinois, Urbana-Champaign\\
{\tt\small \{phuang17,n-ahuja\}@illinois.edu}
\and 
\textsuperscript{2}Facebook \\
{\tt\small \{matzen,jkopf\}@fb.com}
\and 
\textsuperscript{3}Virginia Tech\\
{\tt\small jbhuang@vt.edu}
}

\twocolumn[{
\renewcommand\twocolumn[1][]{#1}
\maketitle
\begin{center}
\hfill
\mpage{0.62}{\includegraphics[width=\linewidth]{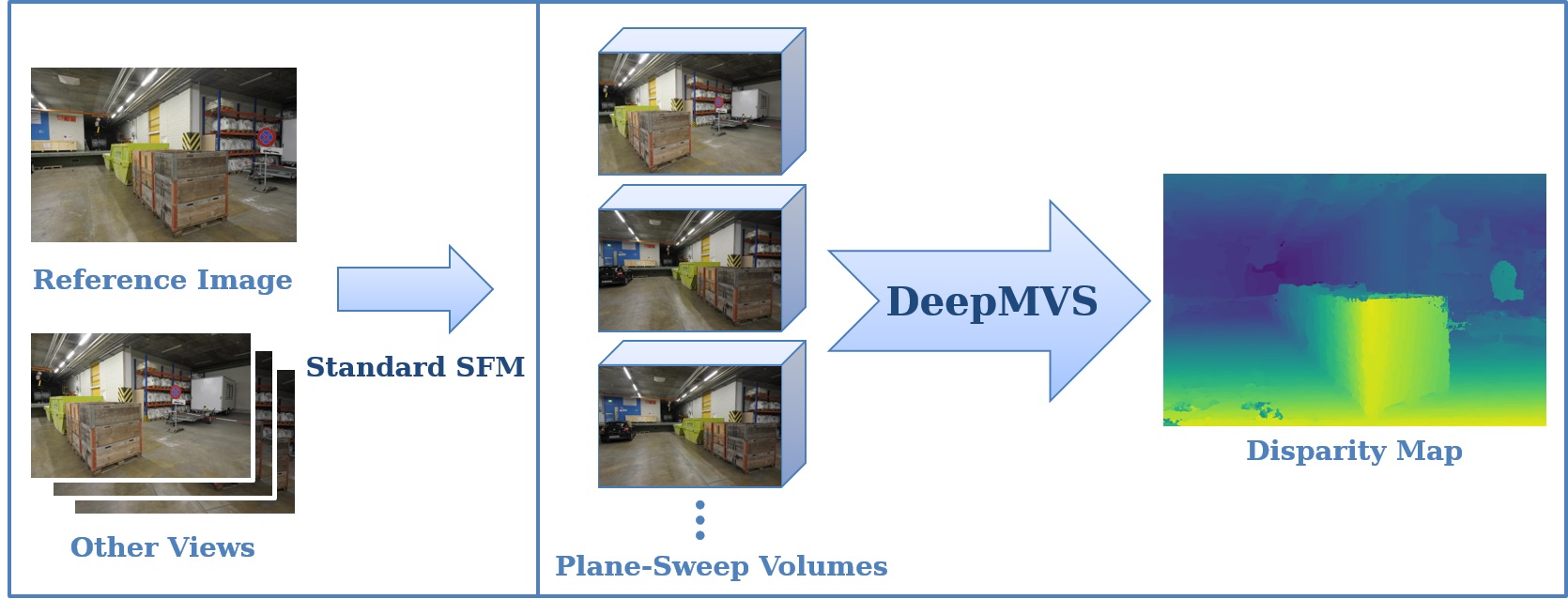}}
\hfill
\mpage{0.36}{
\mpage{0.47}{\includegraphics[width=\linewidth]{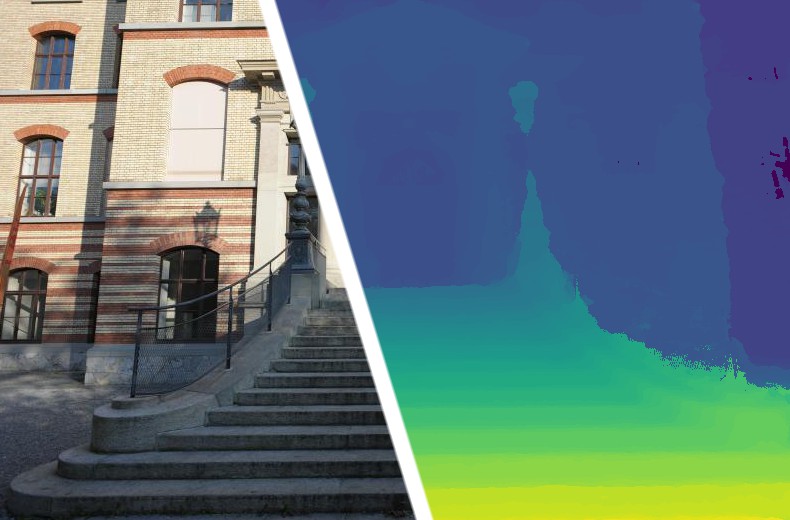}}
\mpage{0.47}{\includegraphics[width=\linewidth]{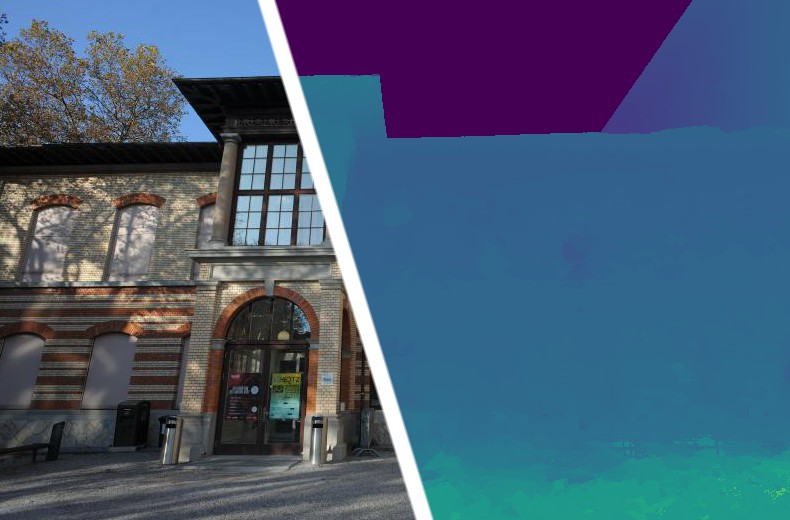}}
\vspace{1mm} \\
\mpage{0.47}{\includegraphics[width=\linewidth]{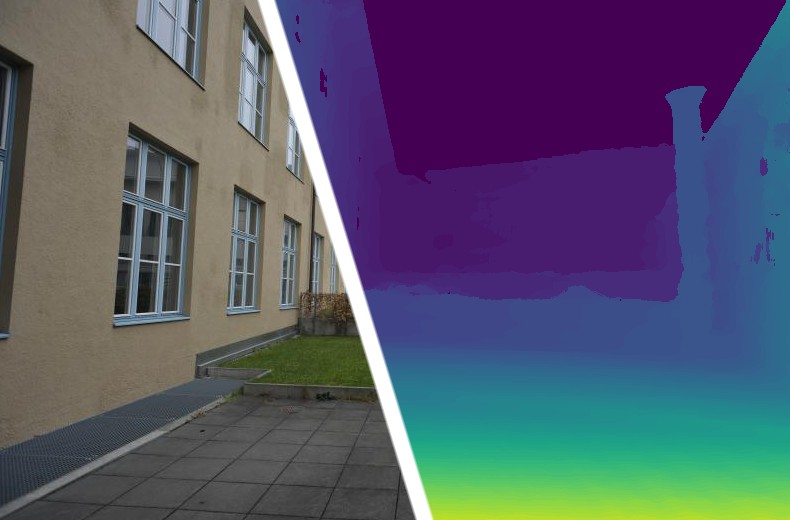}}
\mpage{0.47}{\includegraphics[width=\linewidth]{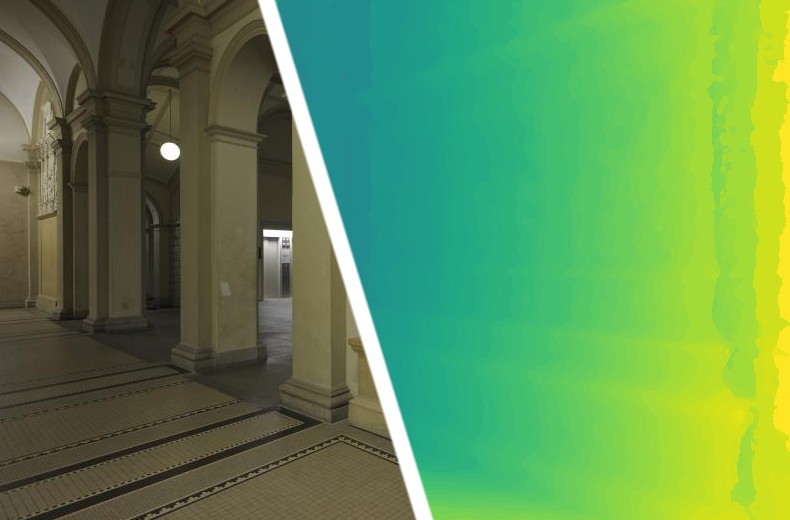}}
}
\vspace{1mm}
\hfill
\mpage{0.16}{(a) Input}\hfill
\mpage{0.45}{(b) Deep multi-view stereo}\hfill
\mpage{0.30}{(c) Sample reconstruction results}
\hfill
\vspace{\figcapmargin}
\captionof{figure}{(a) Our network takes a set of images with known camera poses and calibration as input; 
(b) we produce a set of plane-sweep volumes for a reference view and feed these into a convolutional neural network that predicts a disparity map; 
(c) our network produces high-quality disparity maps even for challenging cases containing poorly textured regions and thin structures.
}
    \label{fig:teaser}
\end{center}



}]

\begin{abstract}

We present DeepMVS, a deep convolutional neural network (ConvNet) for multi-view stereo reconstruction. Taking an arbitrary number of posed images as input, we first produce a set of plane-sweep volumes and use the proposed DeepMVS network to predict high-quality disparity maps. The key contributions that enable these results are (1) supervised pretraining on a photorealistic synthetic dataset, (2) an effective method for aggregating information across a set of unordered images, and (3) integrating multi-layer feature activations from the pre-trained VGG-19 network. We validate the efficacy of DeepMVS using the ETH3D Benchmark. Our results show that DeepMVS compares favorably against state-of-the-art conventional MVS algorithms and other ConvNet based methods, particularly for near-textureless regions and thin structures.

\end{abstract}

\section{Introduction}
\label{sec:intro}

Multi-view Stereo (MVS) methods aim at reconstructing disparity maps from a collection of images with known camera poses and calibration, possibly estimated using Structure from Motion (SFM) algorithms.\footnote{Throughout this work, we always refer to ``disparities'' rather than ``depths''. Disparities are defined as the reciprocal of depths.} MVS is one of the fundamental computer vision problems that have seen decades of research and it is a core component in numerous important applications, including 3D reconstruction, novel view synthesis, augmented reality, and medical imaging~\cite{MVS-TUTORIAL}.

Conventional MVS algorithms often estimate the disparity map by computing plane-sweep volumes and optimizing photometric consistency with handcrafted error functions to measure similarity between patches~\cite{MVS-TUTORIAL}. Aside from photometric consistency, other 3D cues such as lighting~\cite{Wu-CVPR-2011, Langguth-ECCV-2016}, shadows~\cite{Chandraker-CVPR-2007}, color~\cite{Gallup-CVPR-2010}, geometric structures~\cite{Furukawa-CVPR-2009}, and semantic cues~\cite{DENSE-SEMANTIC} have been incorporated into the MVS pipeline for improving the reconstruction accuracy. However, designing algorithms that make explicit use of all these cues is a non-trivial task. Despite extensive research, the results of state-of-the-art MVS algorithms often still contain numerous artifacts, in particular around poorly textured regions, thin structures, and reflective or transparent surfaces~\cite{MVS-TUTORIAL}.

Deep Convolutional Neural Networks (ConvNets) have shown great success in many visual recognition tasks including image classification~\cite{IMAGENET-CLASSIFICATION} and object detection~\cite{RCNN}, as well as in dense pixel-level prediction tasks such as semantic segmentation~\cite{FCN} and optical flow~\cite{FLOWNET,Flownet2}.

For the use of ConvNets in visual reconstruction problems, early work focuses on learning patch similarity for stereo matching~\cite{zbontar2016stereo,zagoruyko2015learning,luo2016efficient}. More recent work performs stereo reconstruction using end-to-end learning. However, these methods either impose constraints on relative camera poses~\cite{LIGHTFIELD,GCNETWORK} or the number of input images~\cite{DEEPSTEREO,DEMON}, or produce a coarse volumetric reconstruction~\cite{3D-R2N2,MVS-MACHINE}.
%

In this paper, we present \textit{DeepMVS}, a deep ConvNet for multi-view stereo that addresses these limitations. Given a reference image and an arbitrary number of neighbor views of the scene, we first perform a standard SFM reconstruction to recover the camera calibration and pose for each image. We then produce a disparity map for the reference image in three stages, as illustrated in~\figref{network-main}. First, we generate a plane-sweep volume for each neighbor image that contains the warped neighbor colors at every disparity, and let our network extract features from each patch pair (reference patch vs.~patch in plane-sweep volume).
Second, we use an encoder-decoder architecture with skip connections to aggregate the features across large spatial regions. We incorporate the feature activations from a VGG-Net~\cite{VGG} pretrained on ImageNet to guide our decoder for disparity predictions. Third, we fuse the information extracted by each neighbor image with a max-pooling layer and produce the final disparity prediction. In contrast to Recurrent Network-based approaches~\cite{3D-R2N2,MVS-MACHINE}, the use of max-pooling allows us to process an arbitrary number of unordered input images

Training deep ConvNets for disparity reconstruction requires a large number of ground truth disparity maps. A solution is to train the network on the combination of a large-scale synthetic dataset and a smaller real-world dataset~\cite{mayer2016large}. Synthetic datasets provide dense pixel-wise ground truth labels for training, but they do not reflect the complexity of realistic photometric effects, illumination, and natural image noise. On the other hand, real-world datasets are limited in scale and often do not have labels for regions in which it is difficult to obtain ground-truth data, such as sky and reflective surfaces. To address this issue, we introduce the \textsc{MVS-Synth} dataset --- a set of 120 photorealistic sequences of synthetic urban scenes for learning-based MVS algorithms. We show that the use of a photorealistic synthetic dataset greatly improves the quality of disparity prediction.

We validate the effectiveness of DeepMVS on the recently introduced ETH3D benchmark dataset~\cite{ETH3D}. Our results show that DeepMVS outperforms DeMoN~\cite{DEMON} in the setting of multi-view stereo, and achieves competitive performance with COLMAP~\cite{COLMAP-MVS}, the state-of-the-art among conventional MVS algorithms. In particular, we observe that our network is often able to produce correct disparities in poorly textured regions, such as sky, walls, floors, and desk surfaces, where conventional algorithms fail.

In summary, we make the following contributions:
\begin{itemize}
\item We propose DeepMVS, a novel learning-based method for multi-view stereo.
\item Unlike existing work~\cite{DEEPSTEREO,DEMON,GCNETWORK}, DeepMVS can process an \emph{arbitrary} number of input images. The disparity estimation result is invariant to the order in which the inputs are processed.
\item Through extensive evaluation, we show that the incorporation of semantic features, training on photorealistic synthetic \textsc{MVS-Synth} dataset, and encoder-decoder architecture for aggregating features over large areas all contribute to the improved performance.
\end{itemize}

\section{Related Work}

\Paragraph{Multi-view stereo reconstruction.} Conventional MVS algorithms focus on designing neighbor selection algorithms and photometric error measures. Recent advances include robust neighbor view selection~\cite{Langguth-ECCV-2016}, incorporation of visibility consistency~\cite{PMVS}, and clustering-based techniques for efficient  reconstruction~\cite{goesele2007multi,furukawa2010towards}. Recently, Sch\"{o}nberger~et al.~present a MVS system --- COLMAP~\cite{COLMAP-MVS} --- that jointly estimates depth and surface normal, leverages photometric and geometric priors for pixelwise view selection, and uses geometric consistency for simultaneous refinement. Through a tight integration of multiple techniques, COLMAP performs among the best algorithms in several public multi-view stereo benchmarks. We refer readers to \cite{MVS-TUTORIAL} for a comprehensive overview of multi-view stereo reconstruction algorithms.

While impressive results have been shown, conventional MVS algorithms rely heavily on photometric consistency and often have difficulty in handling poorly textured and reflective surfaces where photometric consistency is unreliable. In addition, these algorithms do not consider other visual cues for depth perceptions such as lighting, shadows, and semantics (e.g., a building has a planar structure). Incorporating such information through hand-crafted objective functions is non-trivial. In this work, we aim at implicitly leveraging these cues through learning from data.

\Paragraph{Learning-based MVS.}

A line of work focuses on learning a good similarity measure for patch matching across two views~\cite{zbontar2016stereo} and multiple views~\cite{LEARNED-PATCH-MATCHING} using ConvNets.
With the learned stereo matching cost, these methods produce disparity maps by a series of post-processing steps.
In contrast, DeepMVS produces disparity maps directly from a set of posed images.

Another line of recent work uses ConvNets that take a plane-sweep volume as input and produce disparity maps (or synthesizes novel view) for the reference images. However, these approaches assume a fixed number of input images~\cite{GCNETWORK,DEEPSTEREO,DEMON}.
Our proposed DeepMVS can take an \emph{arbitrary} number of images to produce high-quality disparity maps.
Several recent works approach multi-view reconstruction with volumetric methods~\cite{3D-R2N2,MVS-MACHINE}. These methods take a sequence of images captured from different views and generate a 3D shape of the object using a voxel occupancy grid. Nevertheless, the dimension of the voxel grid is quite constrained by the available GPU memory (e.g., coarse grids of 32$\times$32$\times$32 voxels). It is unclear how the volumetric algorithms can be generalized and applied to high-resolution stereo reconstruction in the real-world.

\Paragraph{Learning from simulation.}
Synthetic datasets alleviate the difficulty and the cost of collecting large-scale training datasets from the real world. Examples of synthetic datasets for training and evaluating computer vision algorithms include indoor scene understanding~\cite{ADOBE-INDOOR}, semantic segmentation~\cite{SYNTHIA,PLAYING-FOR-DATA,richter2017playing}, and depth and flow estimation~\cite{mayer2016large,richter2017playing}. We also found that training with a synthetic dataset improves performance in our context. Our newly collected \textsc{MVS-Synth} dataset complements the missing ground truth depth measurements in the real-world such as sky and reflective surfaces like windows.

\section{Learning Multi-view Stereopsis}


The entire pipeline of our algorithm can be broken into four steps. We first preprocess the input image sequence (\secref{Input}), and then generate plane-sweep volumes (\secref{Volume}). Next, our network estimates disparity maps from the plane-sweep volumes (\secref{Network}), followed by final refinement to improve the results (\secref{Refinement}).

\subsection{Input}\label{sec:Input}
The input to our algorithm is a sequence of images and their camera poses and calibration (if necessary, we use the SFM algorithm in COLMAP~\cite{COLMAP-SFM} to estimate them).
One of the input images is designated as the reference image, for which we seek to obtain a disparity map.

We start by selecting a subset of neighbor images for the reference to be used in the stereopsis using a similar approach to COLMAP~\cite{COLMAP-MVS}. The images which share the most common features with the reference are chosen to be neighbor images. However, unlike COLMAP, we do not discard the neighbor images which have small triangulation angles with the reference, and we do not estimate per-image weights, since we intend to train the network so it automatically determines whether a plane-sweep volume is reliable or not by comparing it with the reference image.


We also estimate the disparity range of the reference image. Following the approach as COLMAP, we estimate the maximum disparity by projecting all the features in the sparse reconstruction model to the reference view and computing the disparities of the features.


\subsection{Plane-sweep Volume Generation}\label{sec:Volume}

For each neighbor image we compute a plane-sweep volume with respect to the reference image as follows. We assume that the scene geometry is an infinite plane, fronto-parallel to the reference view, and at specific disparities: $\{ 0, \delta, 2\delta, \ldots, (D-1)\delta\}$. The disparity step, $\delta$, is chosen such that $(D-1)\delta$ equals to the estimated maximum disparity of the reference image.
We warp the neighbor image accordingly and store the result as a layer in the volume.
If any of the assumed disparity is correct and that portion of the scene is not occluded in the neighbor image, we expect that the warped neighbor image matches the reference image well.

By doing this with all the neighbor images, we obtain a stack of plane-sweep volumes with $N \times D$ images, which we denote as $\mathcal{V} = \left\{ V_{n,d}: 0 \le n < N, 0 \le d < D \right\}$. We normalize the RGB values to the range $[-0.5, 0.5]$ and fill the parts in the plane-sweep volumes that are not visible to the corresponding neighbor image with zeros.

The number of disparity levels, $D$, is predetermined. Increasing $D$ allows us to use a smaller disparity step $\delta$ to reduce the quantization errors in the results, but also increases the number of parameters in the network and thus the GPU memory. As a compromise, we choose disparity level $D=100$.

\begin{figure*}[t]
\centering
\includegraphics[width=\textwidth]{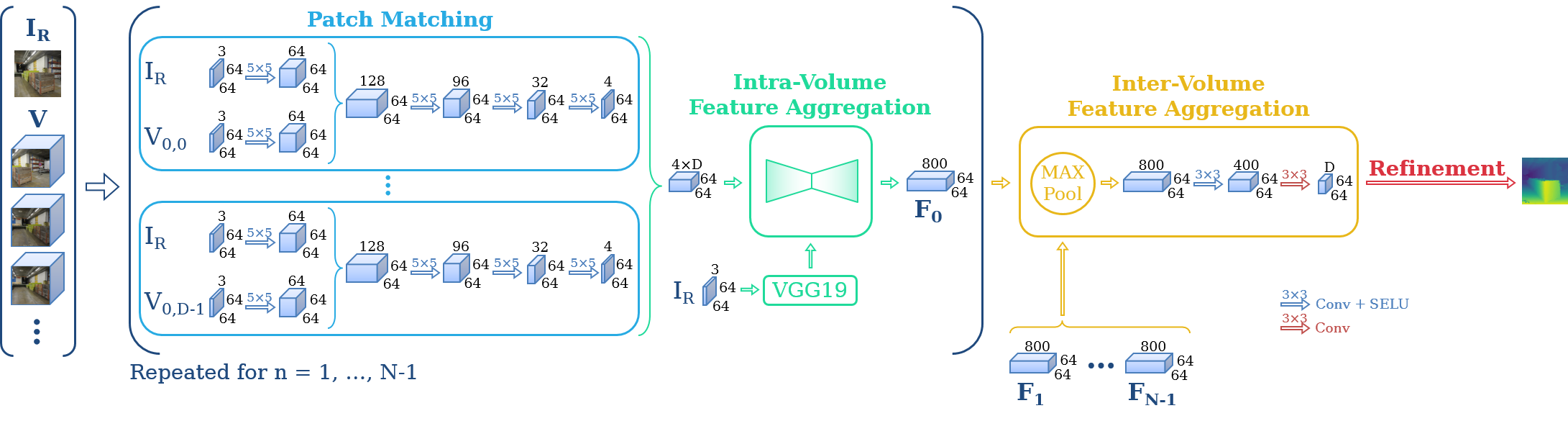}
\caption{DeepMVS network architecture. 
%
}
\label{fig:network-main}
\end{figure*}
\begin{figure*}[t]
\centering
\includegraphics[width=\textwidth]{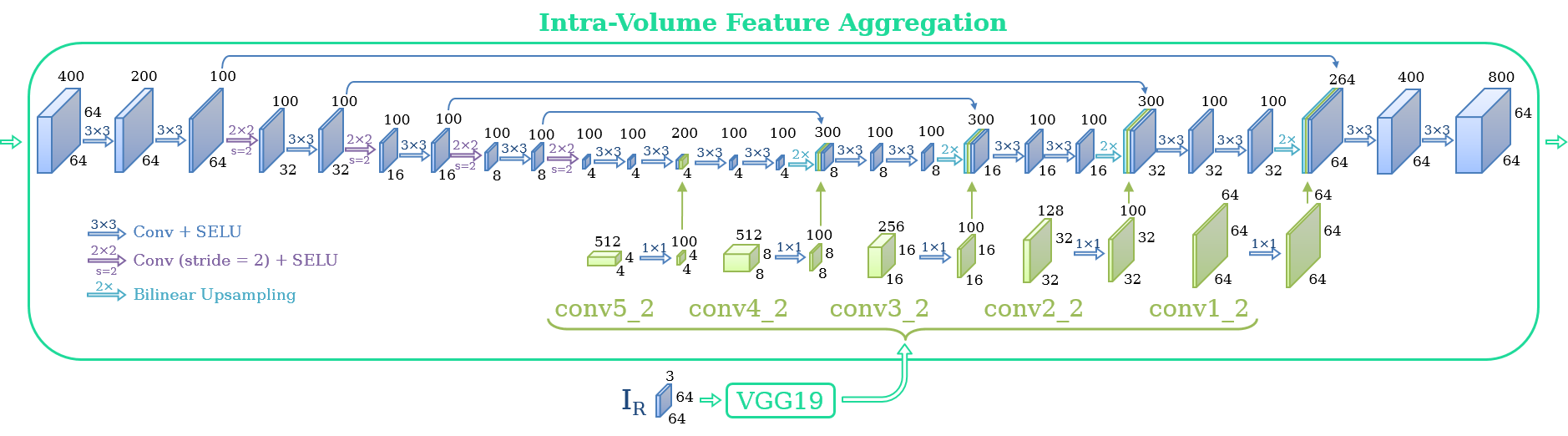}
\caption{Network architecture of the intra-volume feature aggregation network.
}
\label{fig:network-unet}
\end{figure*}

\subsection{Network Architecture}\label{sec:Network}
\figref{network-main} and \figref{network-unet} illustrate the architecture of DeepMVS with the hyper-parameters. Our network can be broken into three parts: 1) the patch matching network, 2) the intra-volume feature aggregation network, and 3) the inter-volume feature aggregation network. Except for the very last layer of the network, all the convolutional layers in the network are followed by a Scaled Exponential Linear Unit (SELU) layer~\cite{SELU}.

\paragraph*{Patch matching.}
The goal of our patch matching network is to extract a set of per-pixel features that can better aid in the comparison of patches than hand-crafted photometric descriptors could do alone.
The patch matching network takes a patch from the reference image $I_R$ and a single patch $V_{n,d}$ from the plane-sweep volume that corresponds to the $n$-th neighbor image at $d$-th disparity level as input. The first convolutional layer extracts 64-channel features from the two patches. The features are then concatenated and passed through three more convolutional layers before turning into 4-channel patch matching features.
We repeat this process for all $N \times D$ plane-swept images.

\paragraph*{Intra-volume feature aggregation.}
For each neighbor image, we concatenate the 4-channel patch matching features of all $D$ disparity levels to form a $4 \times D$-channel volume. Following that is a U-Net structure composed of an encoder, a decoder, and skip connections. Each level of the encoder is formed by a stride-2 convolutional layer followed by an ordinary convolutional layer; each level of the decoder is formed by two convolutional layers followed by a bilinear upsampling layer. We show in \figref{network-unet} the detailed structures and hyper-parameters of the proposed intra-volume feature aggregation network.

In addition, we add semantic features at each level of the decoder. We pass the reference image into the VGG-19~\cite{VGG} network pre-trained on ImageNet, and take the layers \textit{conv1\textunderscore 2}, \textit{conv2\textunderscore 2}, \textit{conv3\textunderscore 2}, \textit{conv4\textunderscore 2}, and \textit{conv5\textunderscore 2} as semantic features. These semantic features are first multiplied by $0.01$ and passed through a convolutional layer so as to reduce dimensionality and to improve numerical stability. Finally, these feature maps are concatenated to each level of the decoder as shown in \figref{network-unet}.

This part of the network is intended to pass the features to larger spatial regions and enable the network to make predictions with non-local information. It also aids the disparity predictions using the VGG feature inputs. The output of the intra-volume feature aggregation network is an 800-channel volume $F_n$ containing the disparity prediction information gathered from the $n$-th neighbor image.



\paragraph*{Inter-volume feature aggregation.}
In this step, we take the $N$ volumes, $\{F_0, \ldots, F_{N-1}\}$, generated from each of the neighbor images and aggregate them using element-wise max-pooling. The use of max-pooling enables the network to gather information from an arbitrary number of neighbor images, and also ensures that the results are invariant with respect to the order of the neighbor images. This technique was previously used in PointNet~\cite{POINTNET} and in the work by Hartmann~\etal in~\cite{LEARNED-PATCH-MATCHING} to allow inputs with varying sizes. Finally, we use two convolutional layers converting the aggregated volume into the pixel-wise disparity predictions.

During training, we randomly select the number of neighbor images $N$ from $\{1,2,3,4\}$. By varying $N$, the network learns to make use of the max-pooling to collect only the useful information from each neighbor image. Even though $N$ is restricted to be no larger than 4 during training (due to the limited size of the GPU memory), we show that our trained network can be applied to an arbitrary number of neighbor images in~\secref{evaluation}.

\paragraph*{Training loss.}
We pose disparity prediction as a multi-class classification problem, and use the cross-entropy loss to train the network. The predicted disparity map can be made by taking the disparity level at which the predicted probability is the highest for each pixel. Namely, for the output distribution $\mathbf{y} = (y_0, \ldots, y_{D-1})$ of each pixel, the predicted disparity can be chosen by
\[ \hat{d}_{\mathrm{raw}} = \argmax_d y_d. \] We refer to this $\hat{d}_{\mathrm{raw}}$ as the \textit{raw predictions}.

\subsection{Refinement}\label{sec:Refinement}

To further improve the quality of the results, we apply the Fully-Connected Conditional Random Field (DenseCRF)~\cite{DENSECRF} to our raw disparity predictions. The use of DenseCRF encourages the pixels which are spatially close and with similar colors to have closer disparity predictions.

\section{Experimental Results}

\subsection{Datasets}\label{sec:datasets}

\paragraph*{DeMoN datasets.} We train our network with the same datasets as used in DeMoN~\cite{DEMON}. The dataset consists of short sequences ranging from two to tens of images including real-world datasets (\textsc{SUN3D}~\cite{SUN3D}, \textsc{RGB-D SLAM}~\cite{RGBD-SLAM}, \textsc{Citywall} and \textsc{Achteck-Turm}~\cite{MVE}) of outdoor and indoor scenes and a synthesized dataset (\textsc{Scenes11}~\cite{DEMON, SHAPENET}) with random objects flying in the air. As suggested in~\cite{DEMON}, mixing real-world and synthetic datasets is important since each has its own limitations. The ground truth for real-world datasets contains measurement errors, whereas synthesized datasets have unrealistic appearance, and may not be capable of reflecting some characteristics of real imagery, such as illumination, depth of field, and noise. The image resolution of this dataset is $640 \times 480$ pixels.

\begin{figure}[t]
\centering
\mpage{0.48}{\includegraphics[width=\linewidth]{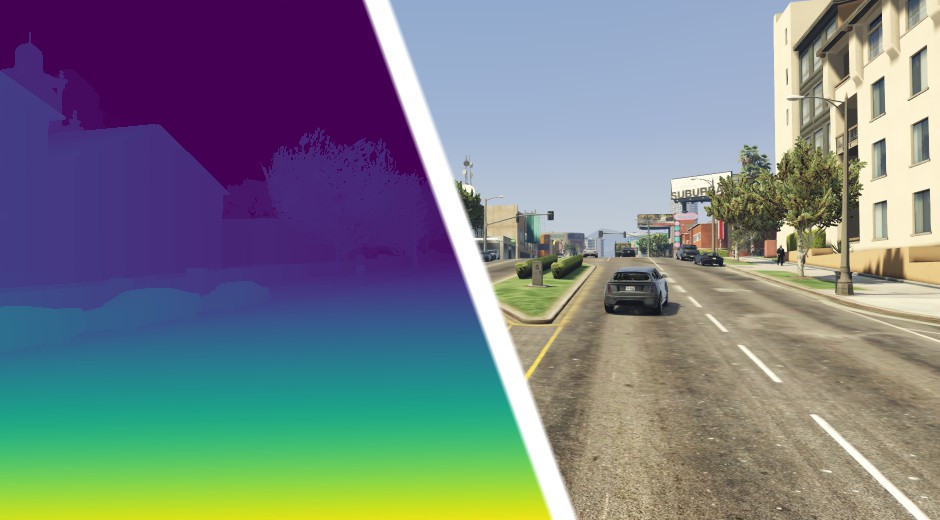}}\hfill
\mpage{0.48}{\includegraphics[width=\linewidth]{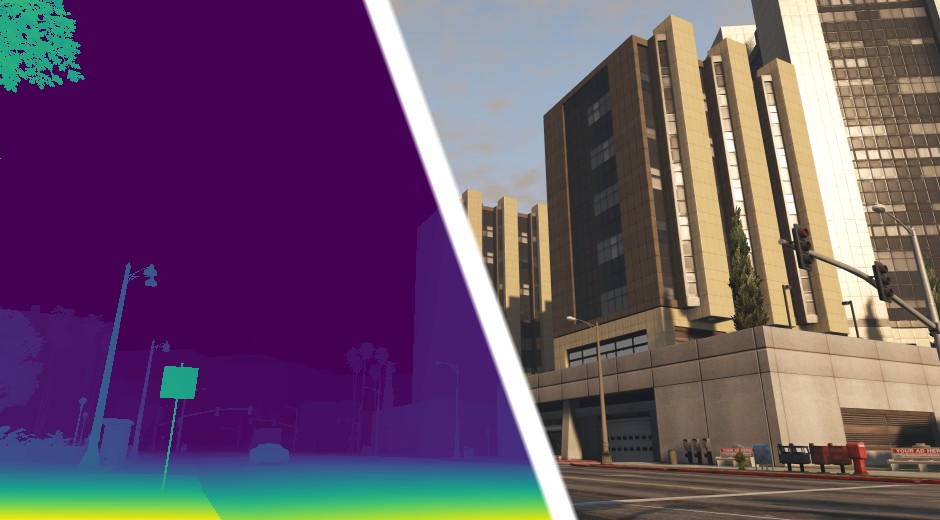}}

\vspace{\figcapmargin}
\caption{
Samples from the proposed \textsc{MVS-Synth} dataset, which provides photorealistic images with ground truth disparities even for the sky, reflective surfaces, and thin structures.
}
\label{fig:GTAV-sample}
\end{figure}

\paragraph*{MVS-Synth dataset.} To address the limitations of the DeMoN datasets, we introduce the \textsc{MVS-Synth} dataset, which consists of 120 sequences of urban scenes captured in the video game \textit{Grand Theft Auto~V}.\footnote{This academic article may contain images and/or data from sources that are not affiliated with the article submitter. Inclusion should not be construed as approval, endorsement or sponsorship of the submitter, article or its content by any such party.} Each sequence is composed of 100 RGB frames of size 1920$\times$1080, ground truth disparity maps, and the extrinsic and intrinsic camera parameters. \figref{GTAV-sample} shows examples from the \textsc{MVS-Synth} dataset.

Compared to existing synthetic datasets, the \textsc{MVS-Synth} dataset is more realistic in terms of context and shading. Compared to real-world datasets, \textsc{MVS-Synth} provides complete ground truth disparities which cover regions such as the sky, reflective surfaces, and thin structures, whose ground truths are usually missing in real-world datasets. Therefore, training with \textsc{MVS-Synth} allows us to predict disparities for these challenging regions. We train the network using both image resolution 1280$\times$720 and 960$\times$540 pixels as data augmentation.

\paragraph*{ETH3D datasets.} For evaluation, we use the high-res multi-view dataset in the recently introduced \textsc{ETH3D} benchmark datasets~\cite{ETH3D}. It consists of 13 sequences of real-world outdoor and indoor scenes with ground truth point clouds captured by laser scanners.
We project the point clouds back to each view to obtain a ground truth disparity map for each reference image. Note that ground truth data are not complete and contain holes in the sky, reflective surfaces, and thin objects. Nevertheless, we use it to validate the efficacy of our method for real-world scenes. We resize the images to 810$\times$540 pixels for evaluation. 


\subsection{Implementation Details}
\label{sec:implementation}
Our training process consists of two stages. First, we train the network by replacing the intra-volume feature aggregation network with two simple 3$\times$3 convolutional layers. Here, our goal in the first stage is to pre-train the network so it can be transferred to the second stage. Then, we add the intra-volume feature aggregation network back with weights initialized from the pre-trained network, and train the entire network using both DeMoN and the \textsc{MVS-Synth} datasets.

For both training stages, we use the Adam solver~\cite{ADAM} with learning rates $10^{-5}$ and $10^{-6}$, respectively, for 320k iterations per stage. We apply gradient clipping to prevent gradient explosion by constraining the L2-norm of the gradients at each layer to be no more than $1.0$ at the first stage and $0.1$ at the second stage. 
We implement the network in PyTorch. Training the network with an NVIDIA P100 GPU with 16GB memory takes two days for each stage.

We use 64px$\times$64px patches as our inputs so as to fit our network into the GPU memory at the  training stages. We generate the semantic features by a feed-forward pass of a VGG-19 network using the entire image. We then take only the region of interest corresponding to the input patches from the intermediate features. At test time, we feed 128px$\times$128px patches into the network, and take only the center 64px$\times$64px of the output to reduce boundary artifacts. The 64px$\times$64px output patches are then tiled to achieve full-resolution results.

\subsection{Evaluation Metrics}

\Paragraph{Geometric errors.} We compute geometric error by taking the L1 distance between the predicted disparity and the ground truth. Unavailable pixels are ignored. 

\Paragraph{Photometric errors.} We also measure photometric \emph{rephotography error}~\cite{WAECHTER} --- the L1 distance between the reference and the rephotography image. We generate the rephotography using the predicted disparity map, warping the pixels to all other neighbor images, sampling colors from the neighbor images, and finally selecting the median among all color candidates for each pixel.

\Paragraph{Completeness.}
Another important factor for evaluation is completeness. We measure completeness using the percentage of pixels whose errors are below a certain threshold. Plotting the relationship between different error thresholds and their corresponding completeness helps visualize the distributions of the errors. The curves lying in the lower right represent more pixels having lower errors and thus have better performance.

\subsection{Evaluation}\label{sec:evaluation}

\Paragraph{COLMAP.}
Several conventional MVS algorithms have been proposed, including PMVS~\cite{PMVS}, MVE~\cite{MVE}, and COLMAP~\cite{COLMAP-MVS}. We choose to compare with COLMAP as it is the top performer on the \textsc{ETH3D} dataset~\cite{ETH3D}.

We follow the default settings of COLMAP unless otherwise mentioned. 
%
COLMAP provides an option to filter out the predictions that are not geometrically consistent. However, the filtered disparity maps may significantly reduce completeness. We show both unfiltered and filtered maps for comparison. 

Note that we do not use DenseCRF to refine COLMAP's noisy unfiltered maps since COLMAP predicts a \emph{deterministic} disparity for each pixel, whereas DenseCRF requires pixel-wise distributions as inputs.

\Paragraph{DeMoN.}
We compare our approach with DeMoN~\cite{DEMON} because it is the closest to ours among the existing learning-based stereopsis methods. However, as their network only works with image \emph{pairs}, we propose two ways to extend their approach to multi-view stereo applications.

The first method is to choose the best result among all the disparity maps generated from the image pairs formed by the reference image and its neighbor images. This method is not practical in real applications since the ground truths are not available. Nevertheless, the method establishes the upper-bound performance of DeMoN. The second method is to compute the per-pixel median among all the generated disparity maps so as to aggregate information from all available image pairs. 

\begin{figure*}[t]
	\centering
    \includegraphics[width=0.135\textwidth]{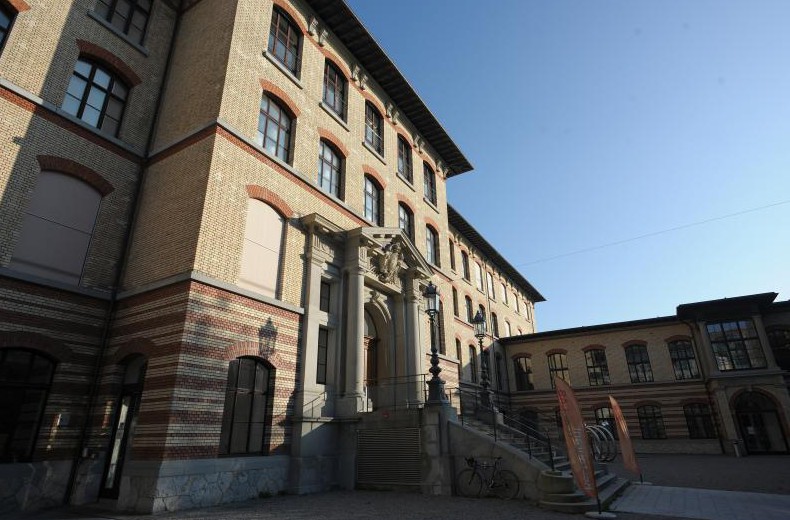}
    \includegraphics[width=0.135\textwidth]{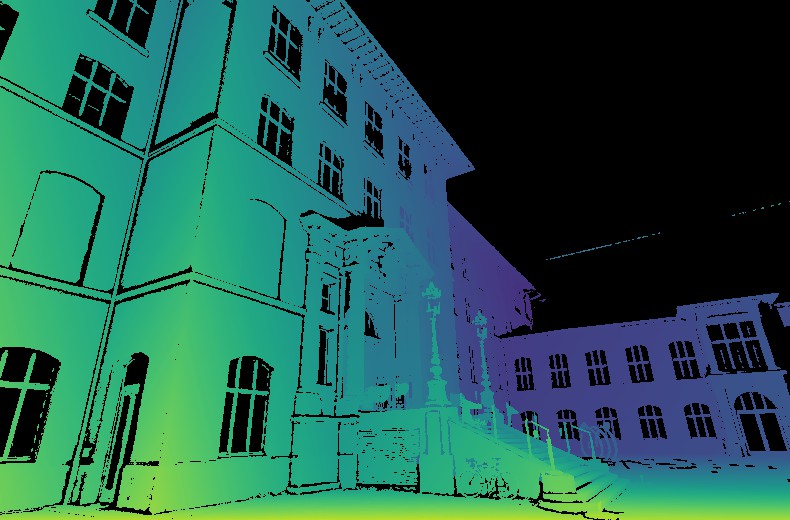}
    \includegraphics[width=0.135\textwidth]{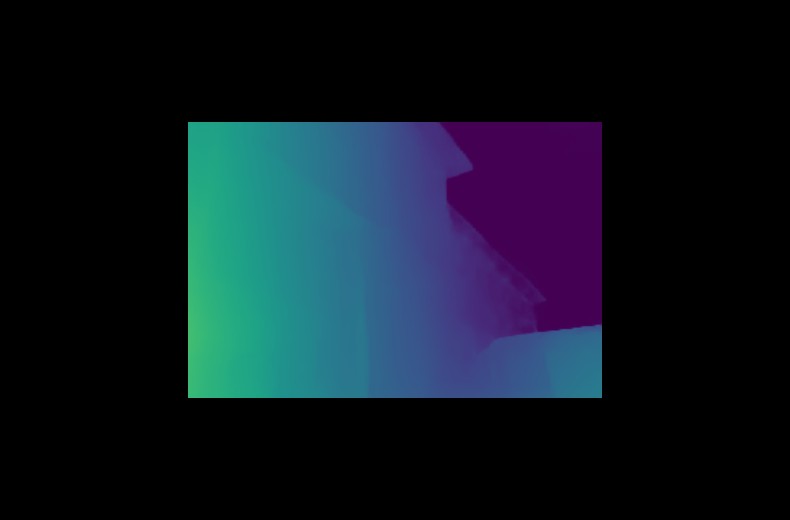}
    \includegraphics[width=0.135\textwidth]{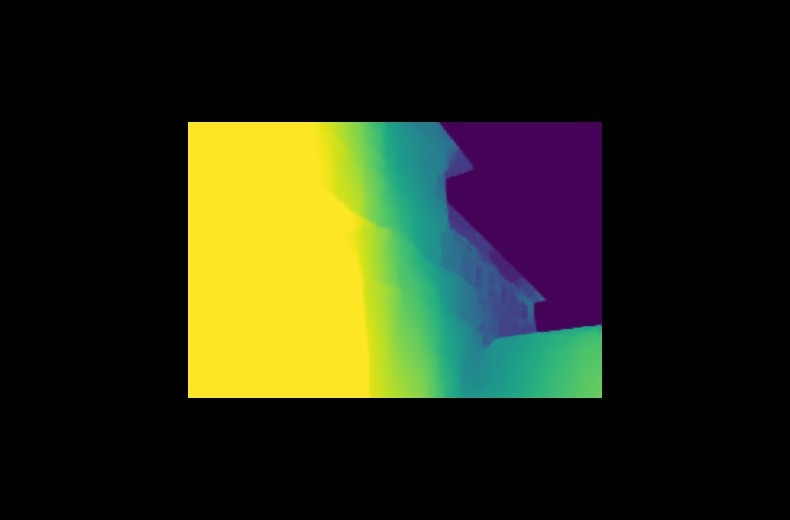}
    \includegraphics[width=0.135\textwidth]{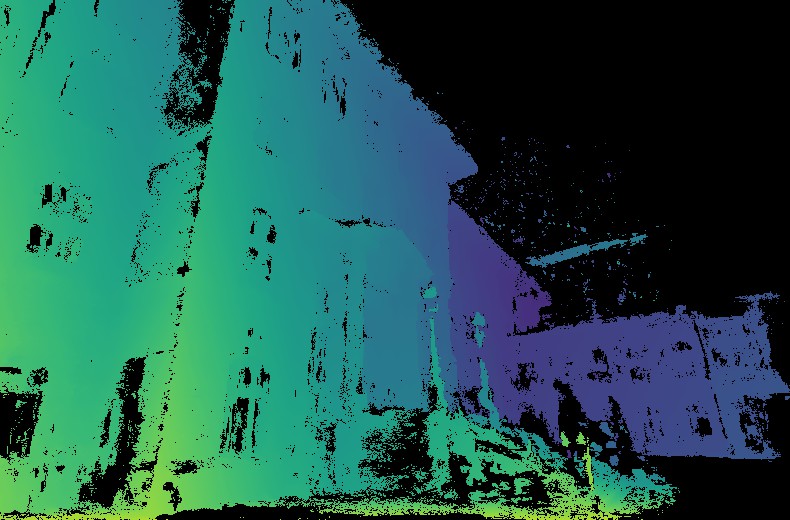}
    \includegraphics[width=0.135\textwidth]{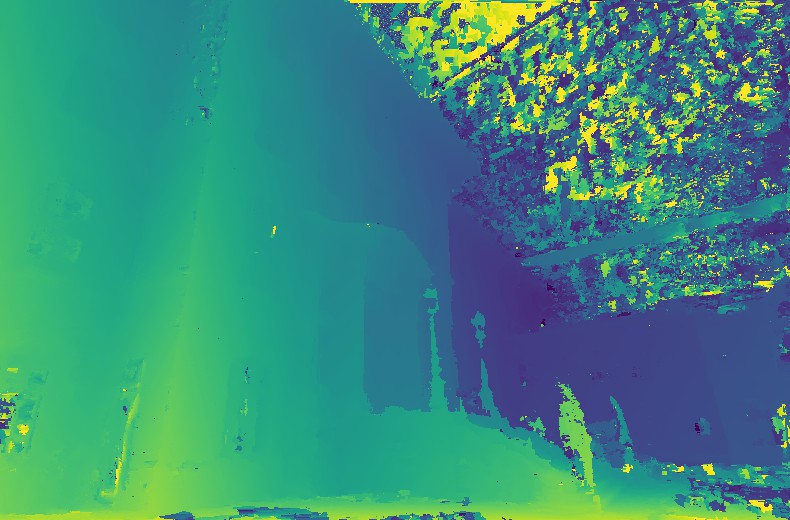}
    \includegraphics[width=0.135\textwidth]{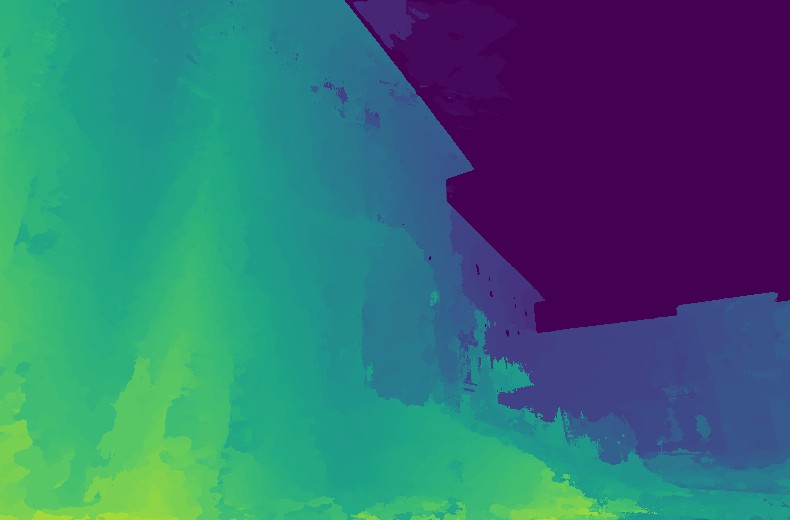} \\
    \vspace{0.5mm}
    \includegraphics[width=0.135\textwidth]{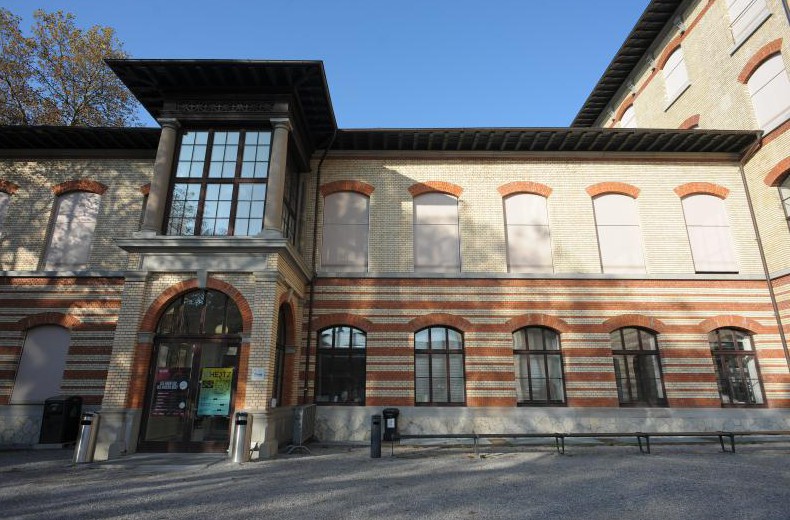}
    \includegraphics[width=0.135\textwidth]{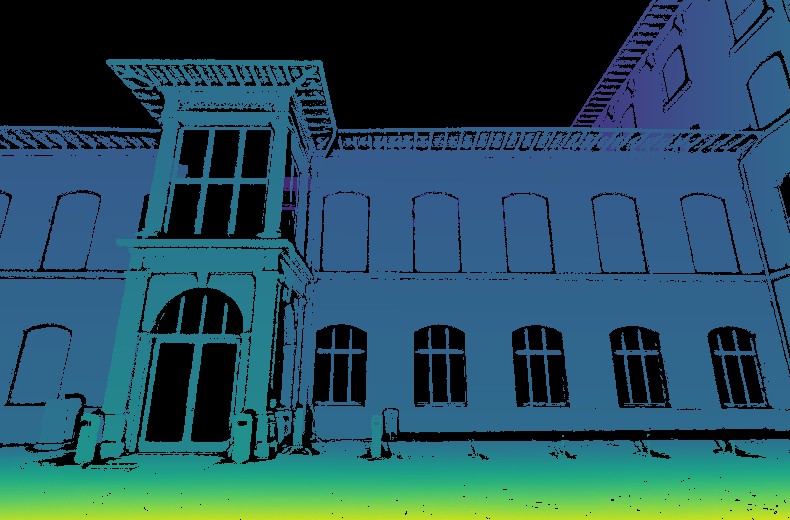}
    \includegraphics[width=0.135\textwidth]{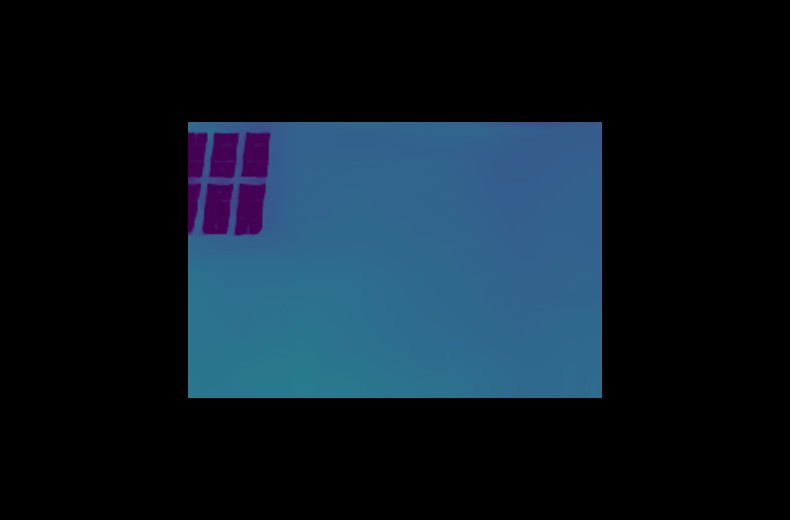}
    \includegraphics[width=0.135\textwidth]{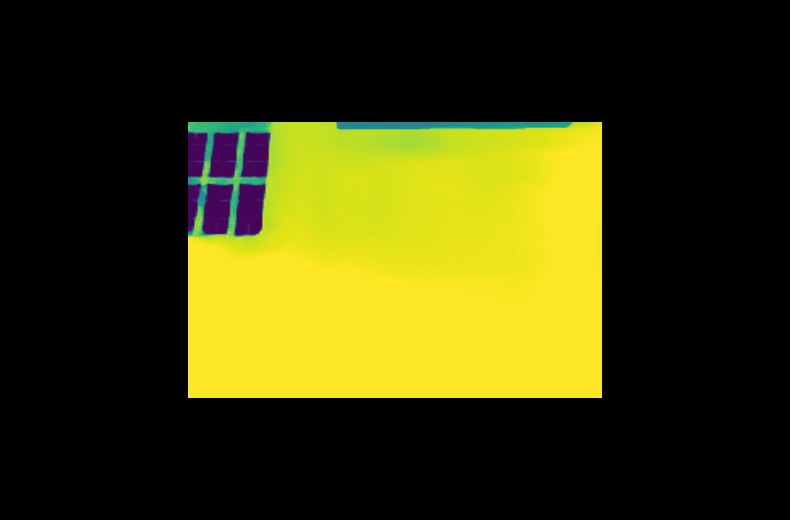}
    \includegraphics[width=0.135\textwidth]{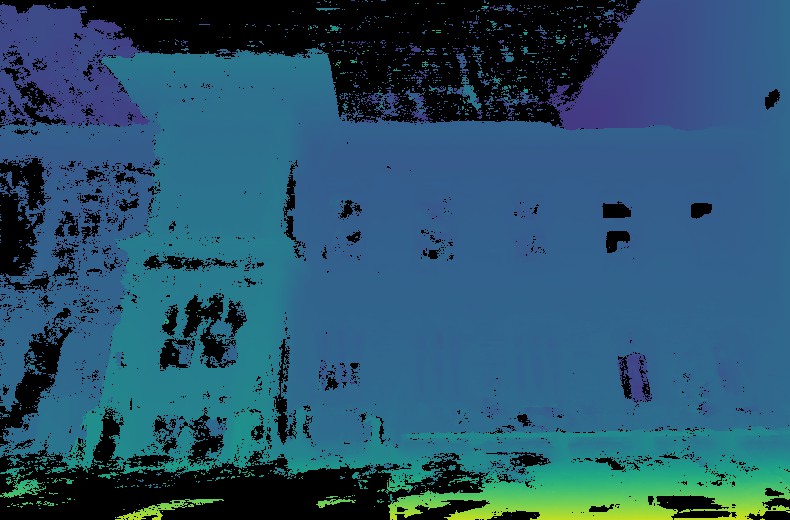}
    \includegraphics[width=0.135\textwidth]{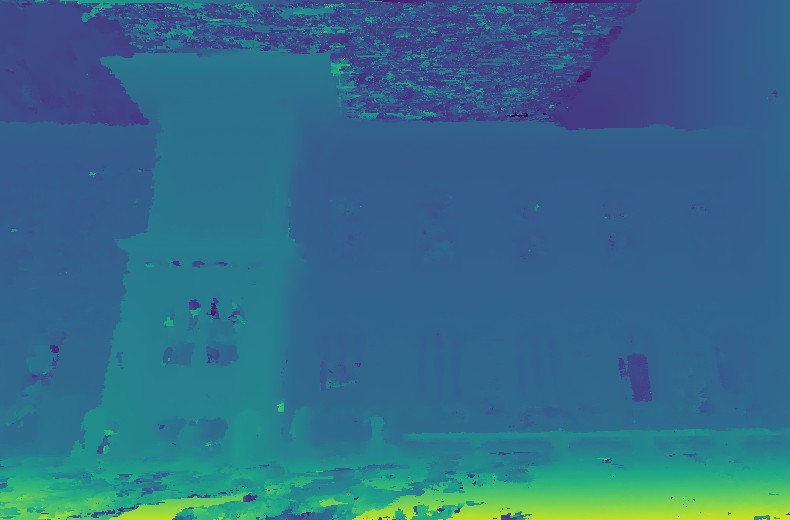}
    \includegraphics[width=0.135\textwidth]{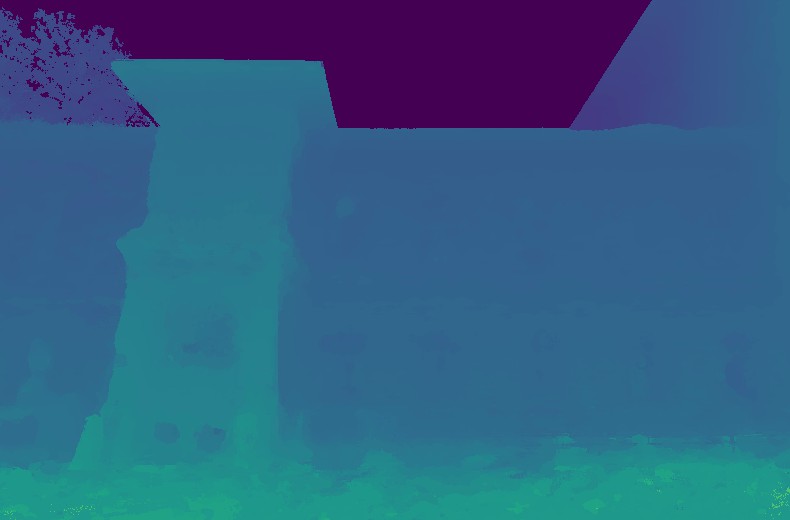} \\
    \vspace{0.5mm}
    \includegraphics[width=0.135\textwidth]{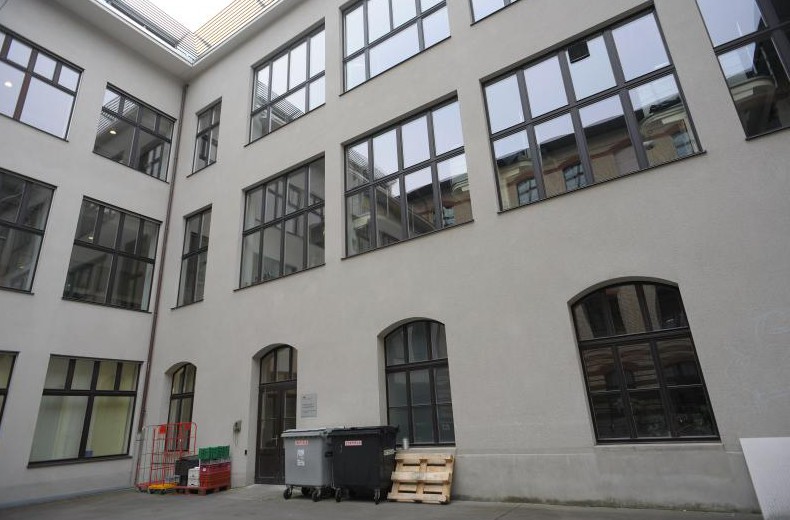}
    \includegraphics[width=0.135\textwidth]{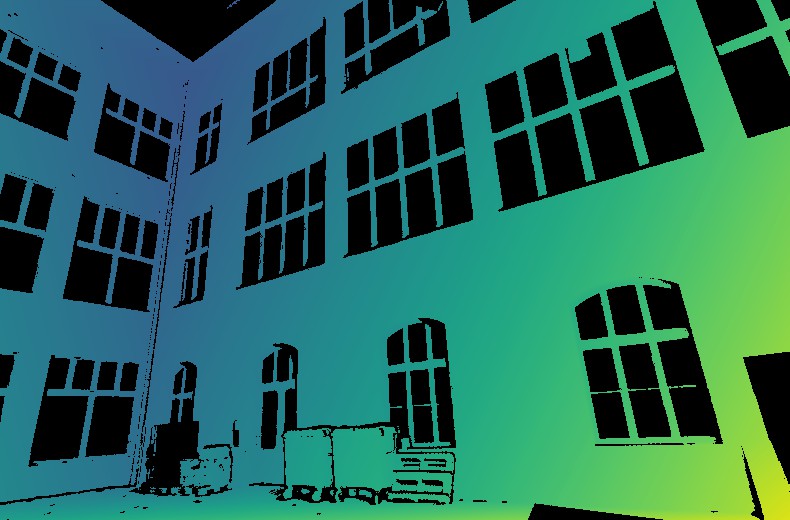}
    \includegraphics[width=0.135\textwidth]{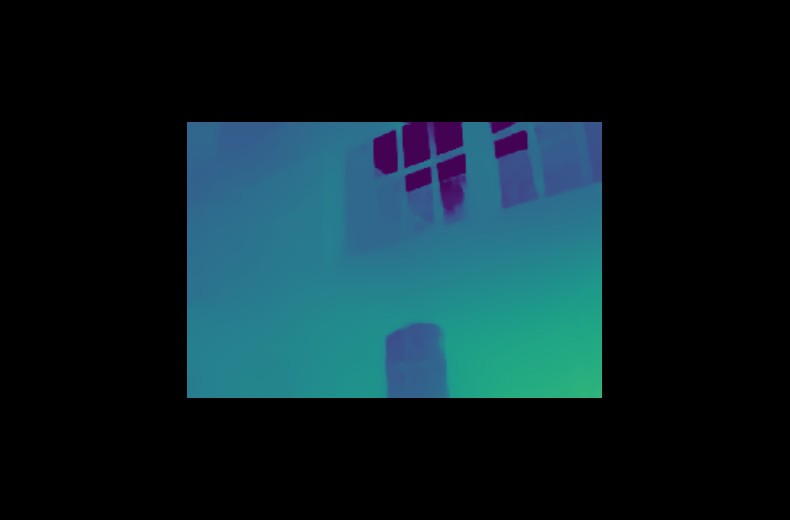}
    \includegraphics[width=0.135\textwidth]{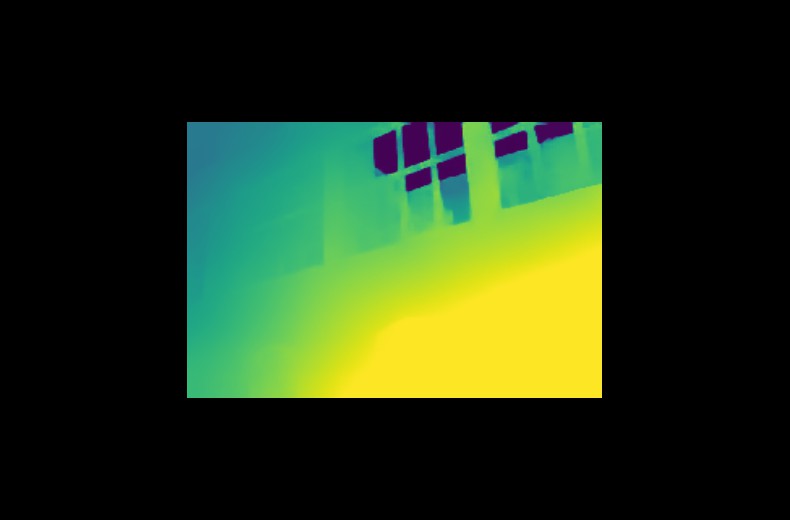}
    \includegraphics[width=0.135\textwidth]{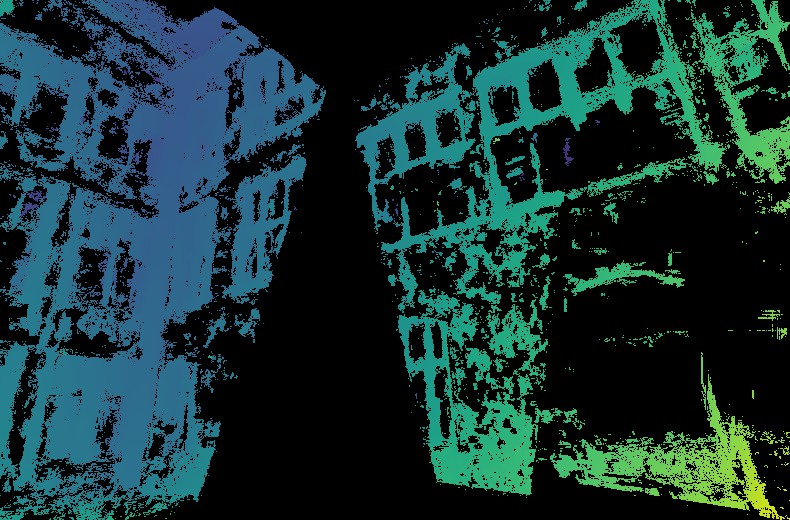}
    \includegraphics[width=0.135\textwidth]{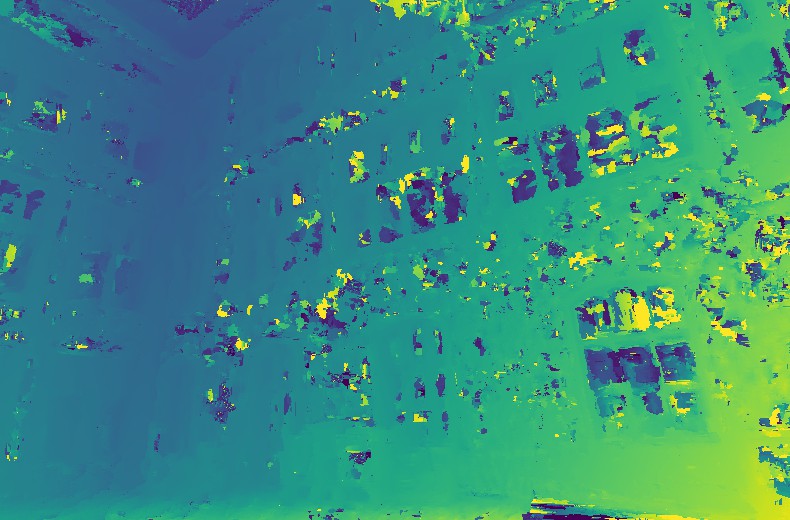}
    \includegraphics[width=0.135\textwidth]{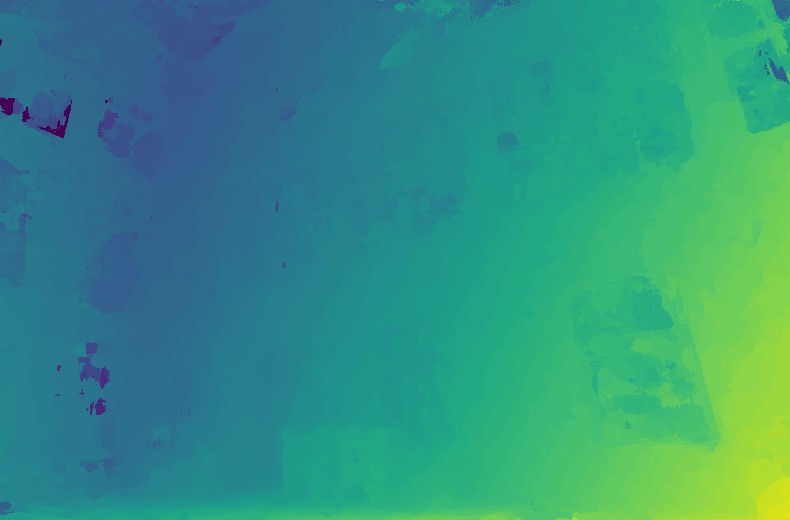} \\
    \vspace{0.5mm}
    \includegraphics[width=0.135\textwidth]{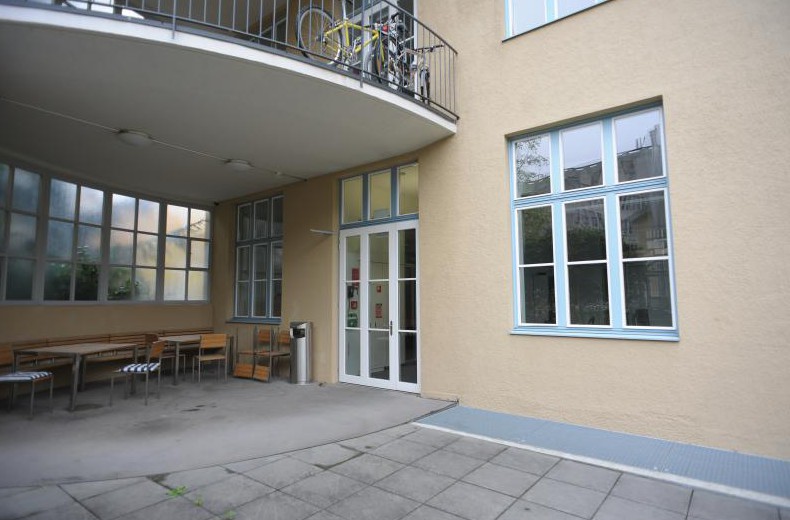}
    \includegraphics[width=0.135\textwidth]{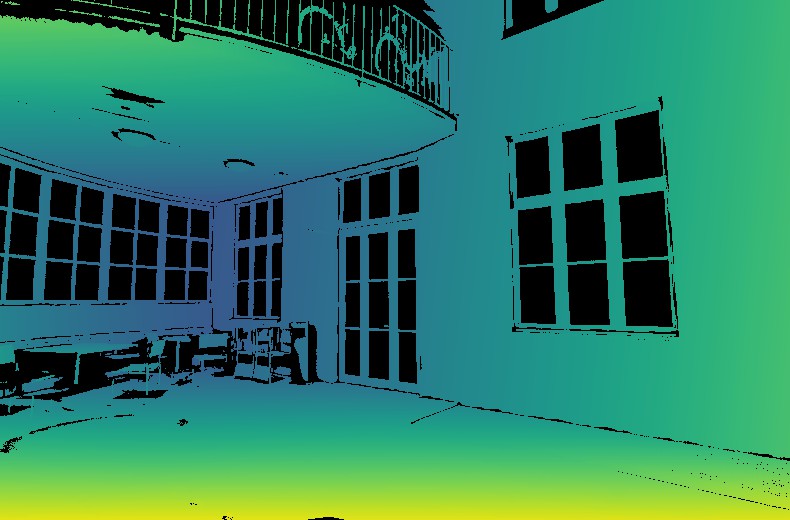}
    \includegraphics[width=0.135\textwidth]{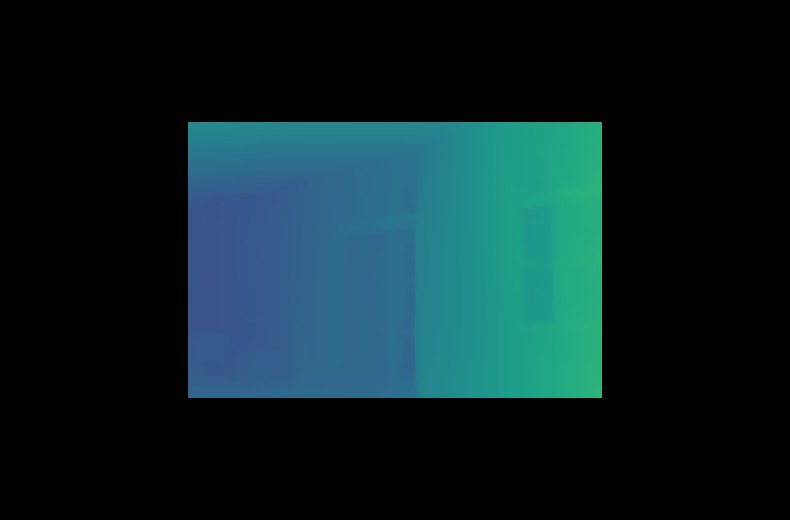}
    \includegraphics[width=0.135\textwidth]{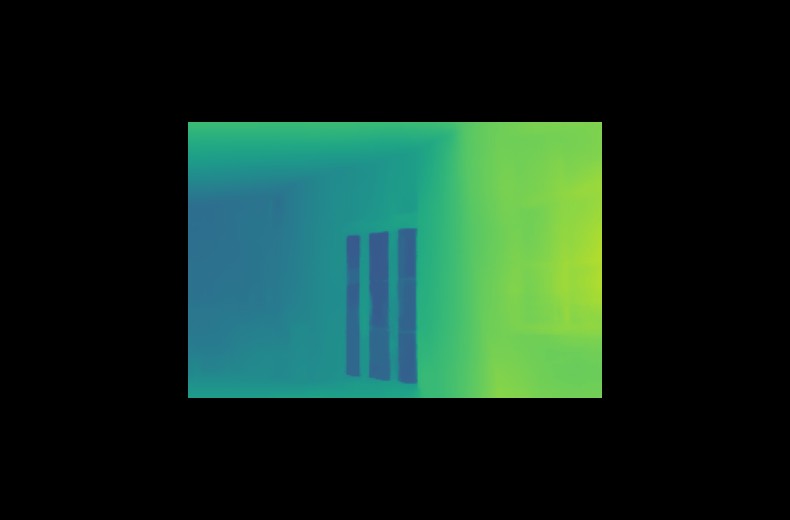}
    \includegraphics[width=0.135\textwidth]{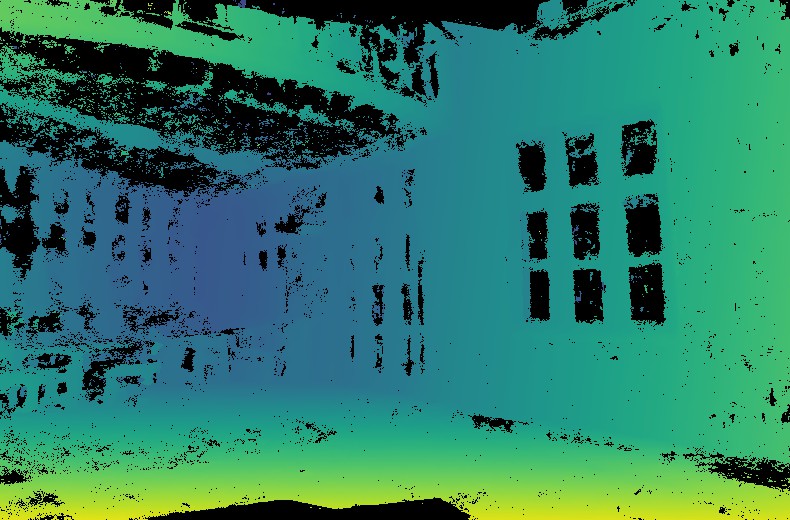}
    \includegraphics[width=0.135\textwidth]{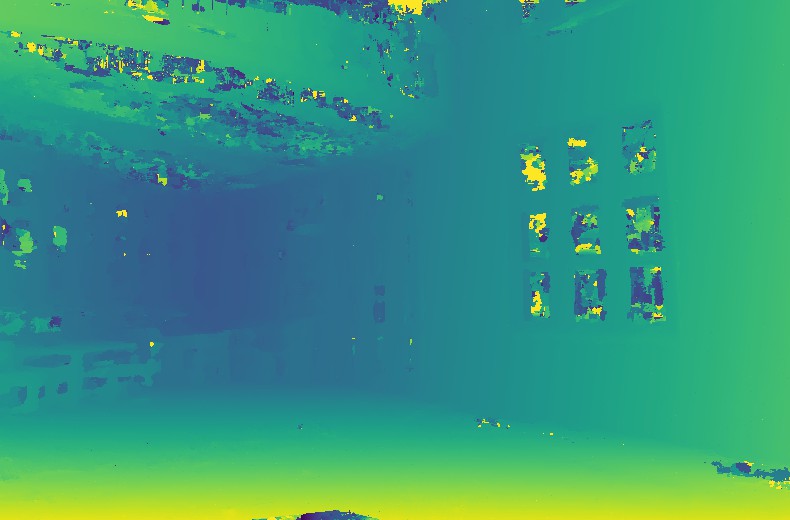}
    \includegraphics[width=0.135\textwidth]{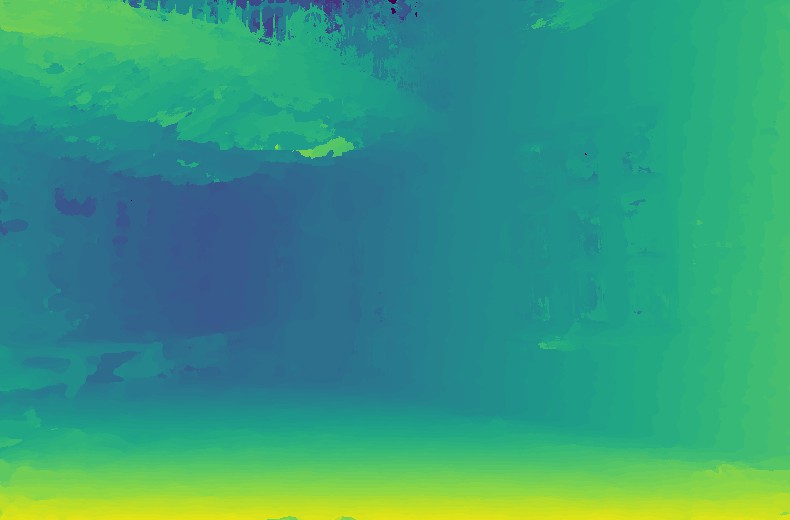} \\
    \vspace{0.5mm}
    \includegraphics[width=0.135\textwidth]{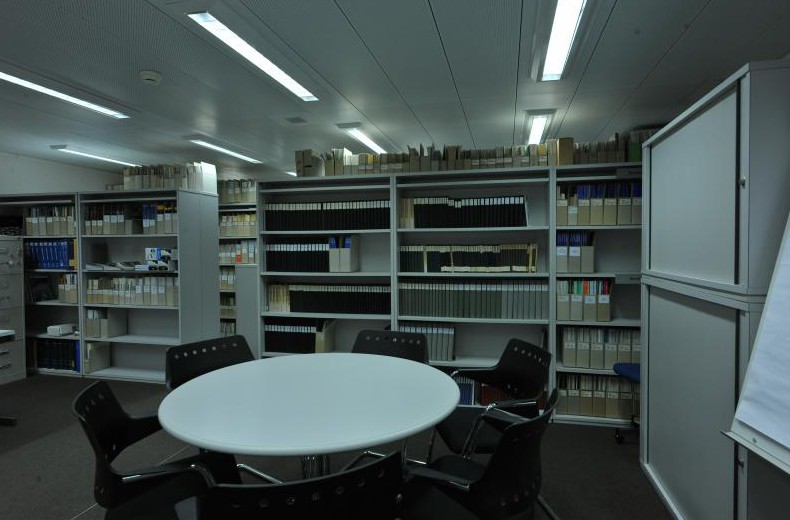}
    \includegraphics[width=0.135\textwidth]{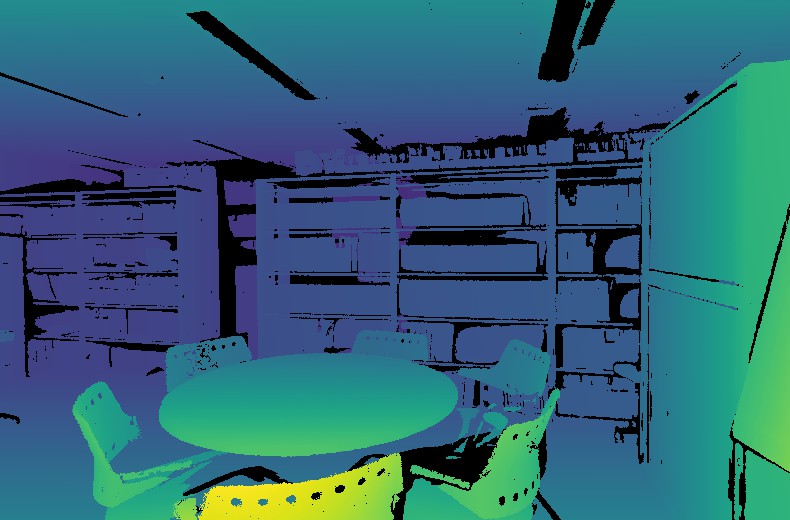}
    \includegraphics[width=0.135\textwidth]{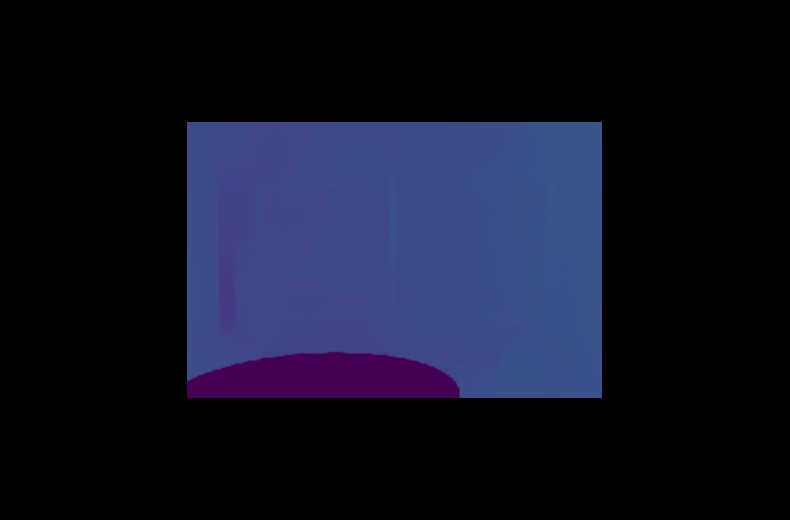}
    \includegraphics[width=0.135\textwidth]{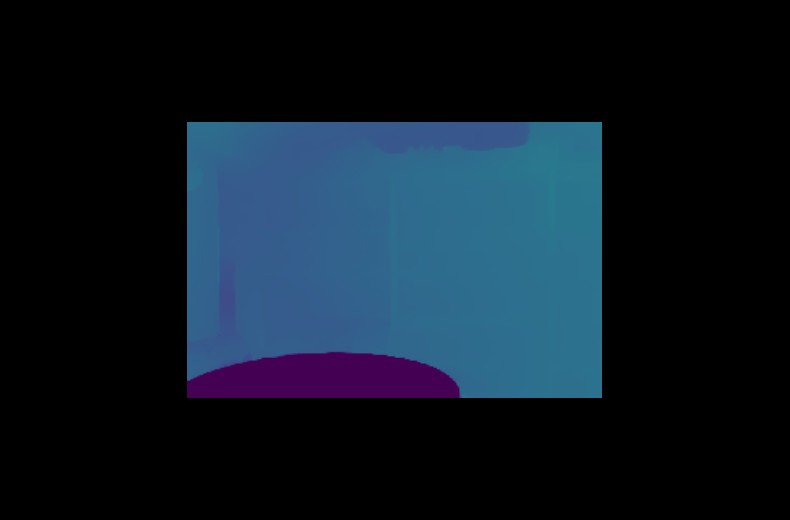}
    \includegraphics[width=0.135\textwidth]{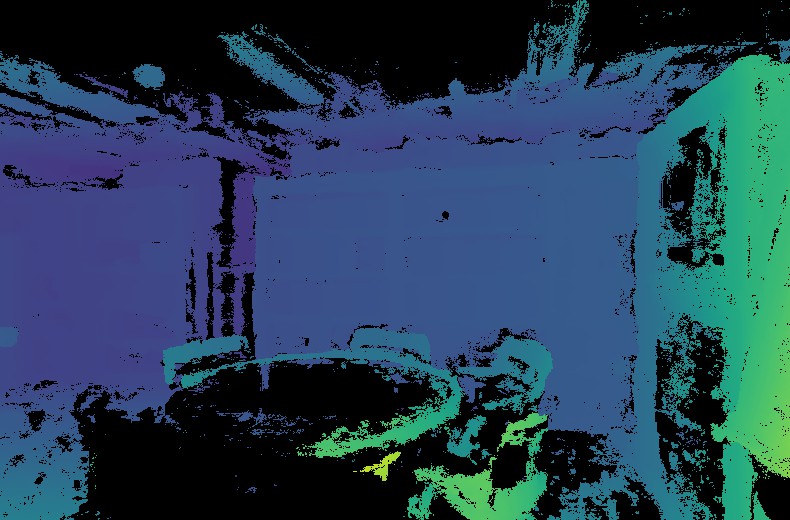}
    \includegraphics[width=0.135\textwidth]{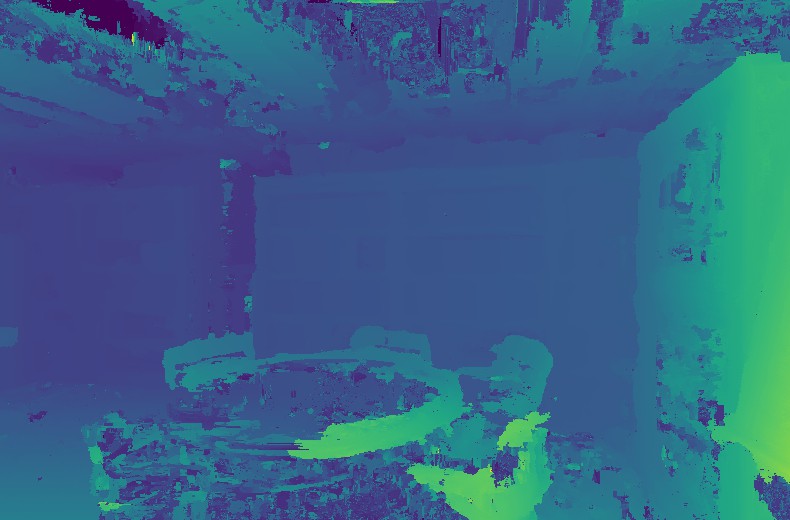}
    \includegraphics[width=0.135\textwidth]{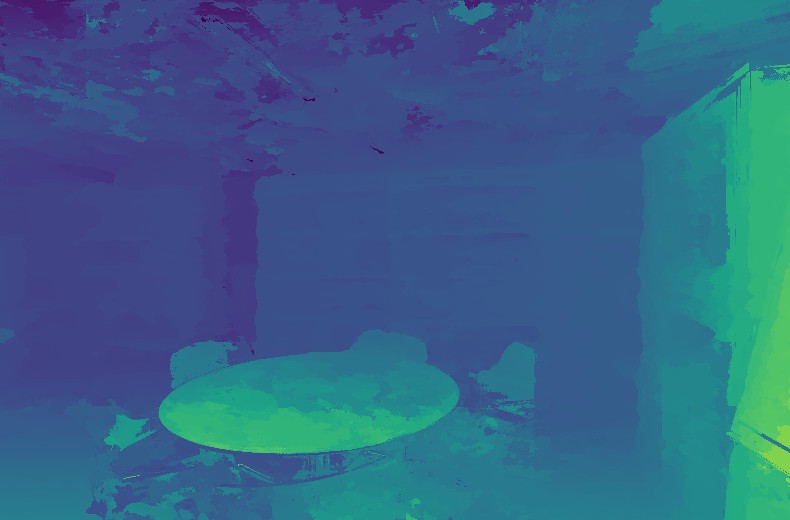} \\
    \vspace{0.5mm}
    \includegraphics[width=0.135\textwidth]{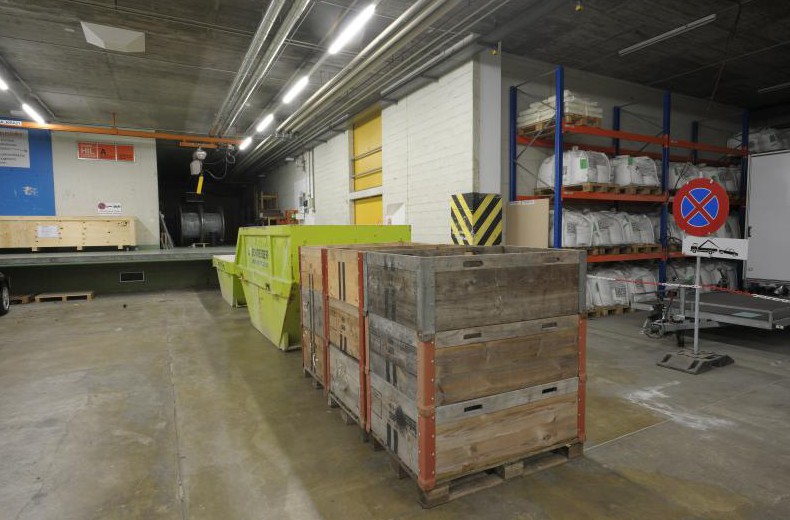}
    \includegraphics[width=0.135\textwidth]{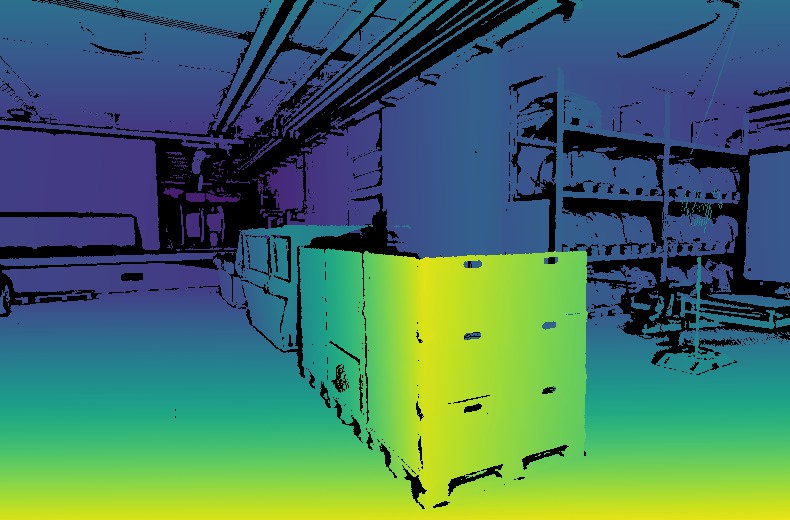}
    \includegraphics[width=0.135\textwidth]{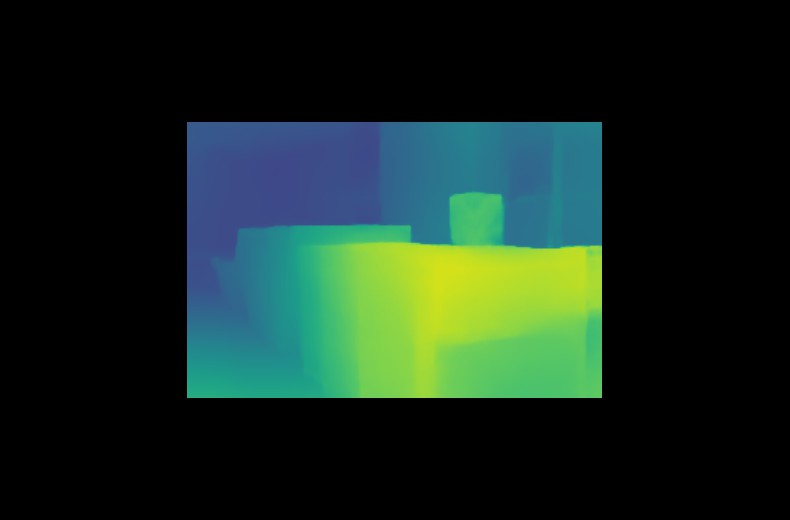}
    \includegraphics[width=0.135\textwidth]{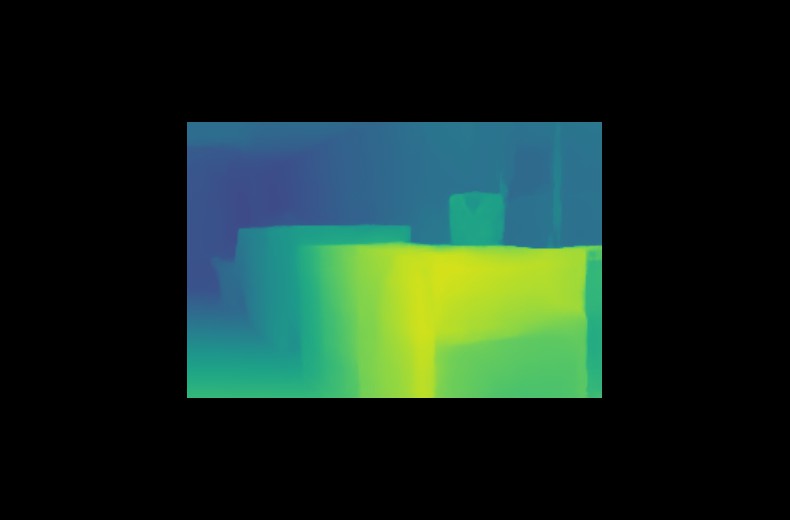}
    \includegraphics[width=0.135\textwidth]{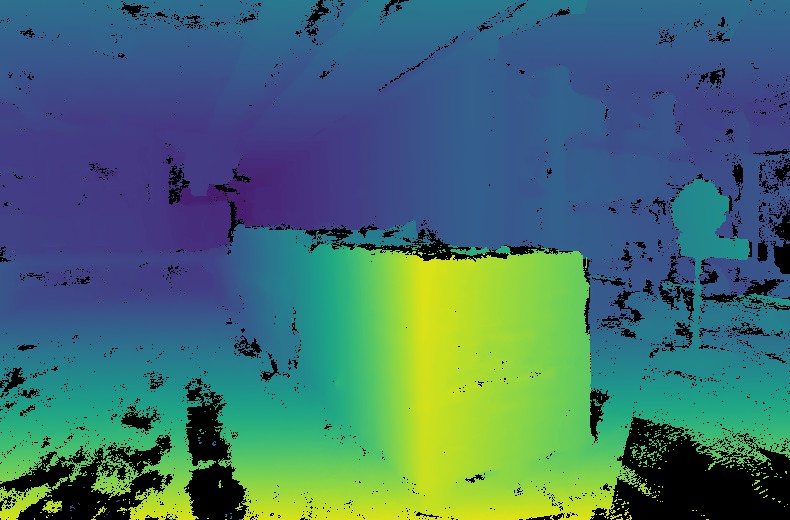}
    \includegraphics[width=0.135\textwidth]{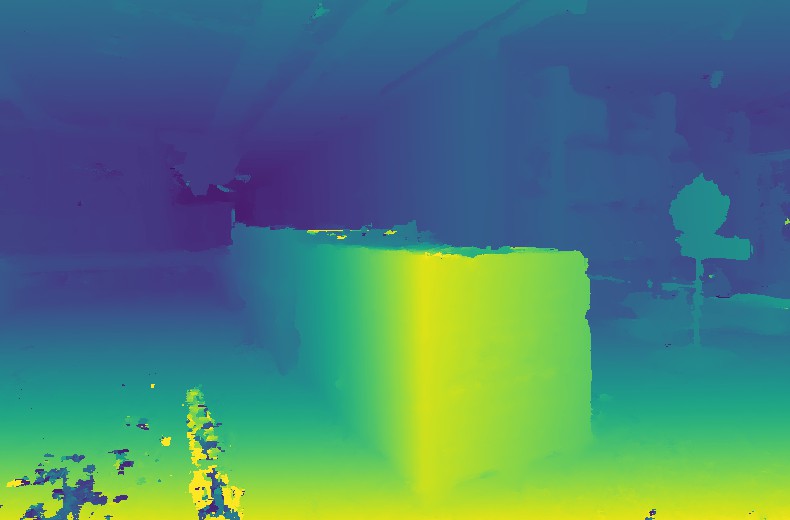}
    \includegraphics[width=0.135\textwidth]{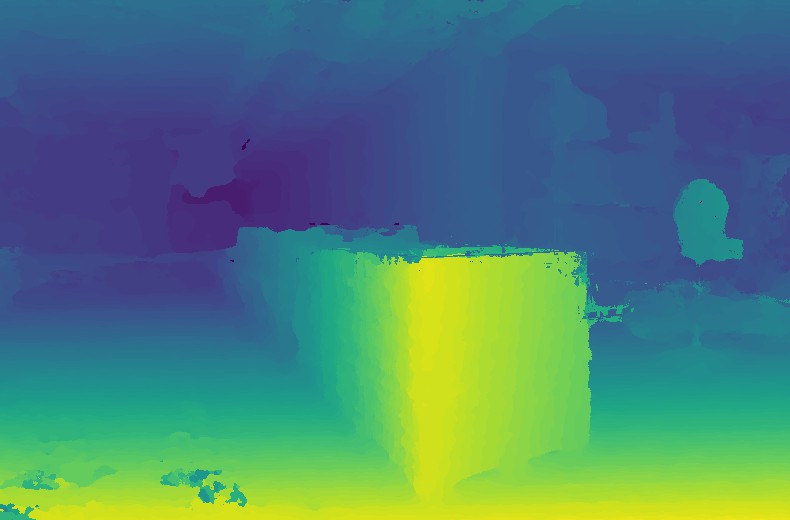} \\
    \vspace{0.5mm}
    \includegraphics[width=0.135\textwidth]{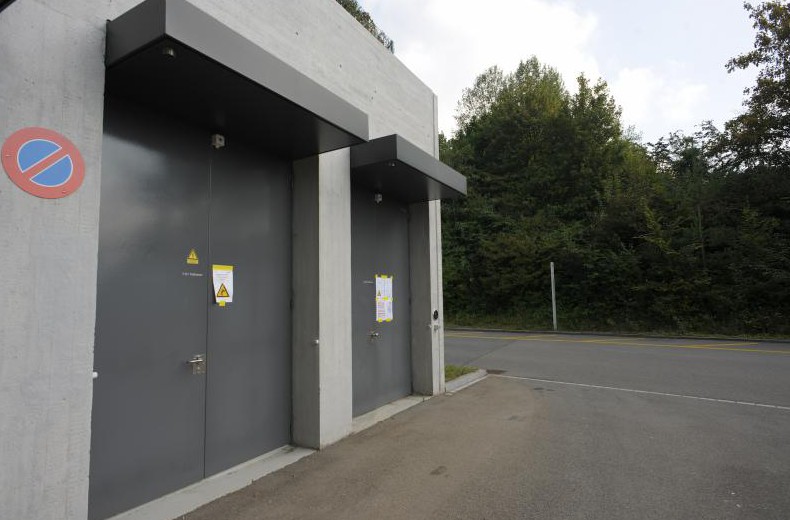}
    \includegraphics[width=0.135\textwidth]{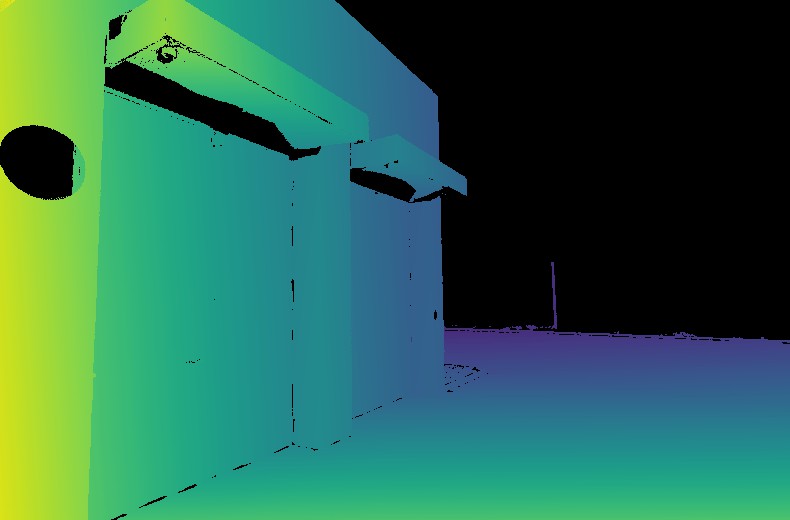}
    \includegraphics[width=0.135\textwidth]{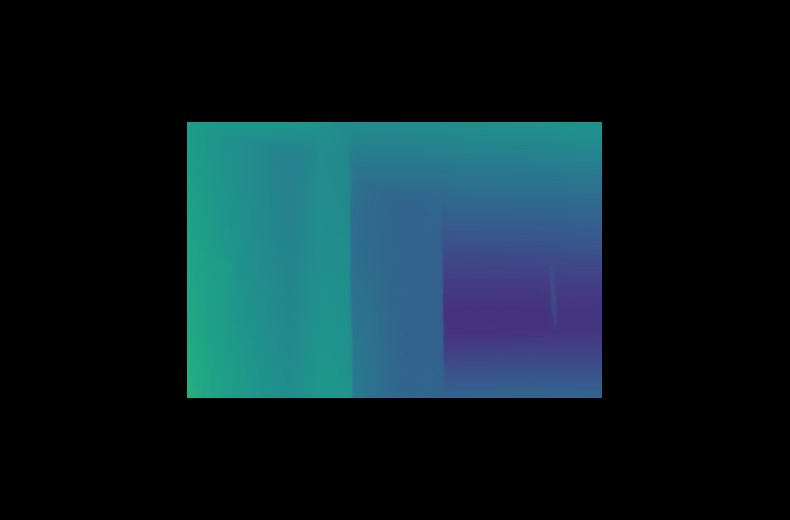}
    \includegraphics[width=0.135\textwidth]{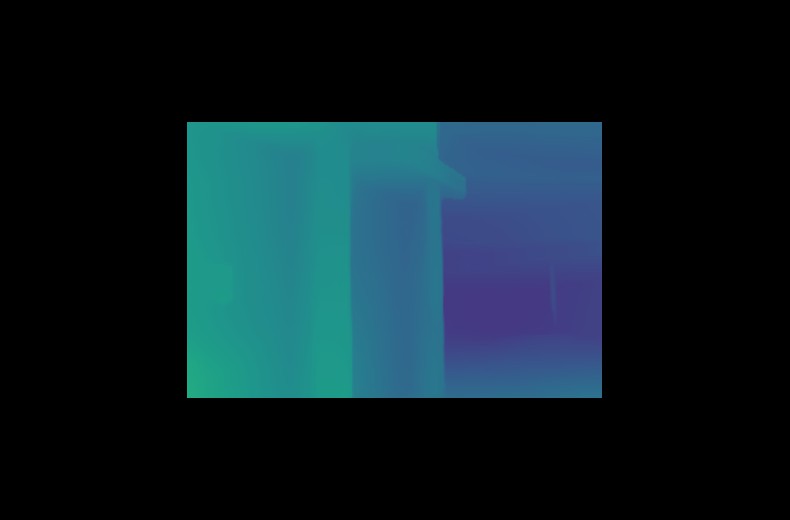}
    \includegraphics[width=0.135\textwidth]{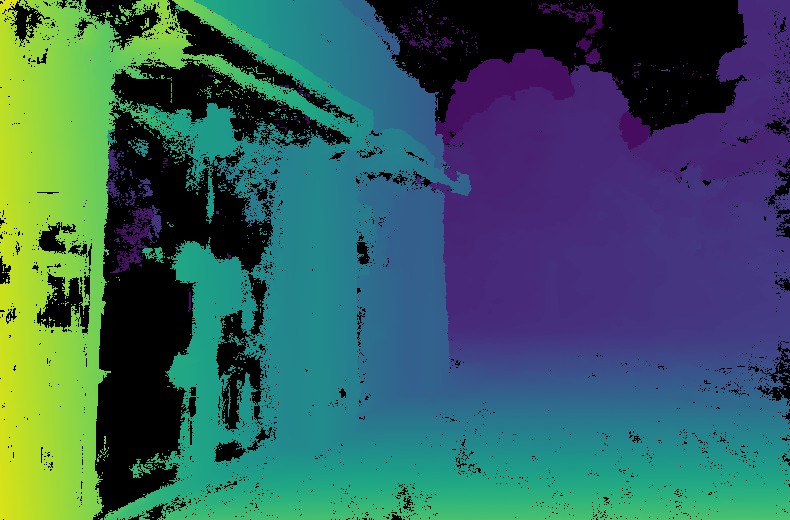}
    \includegraphics[width=0.135\textwidth]{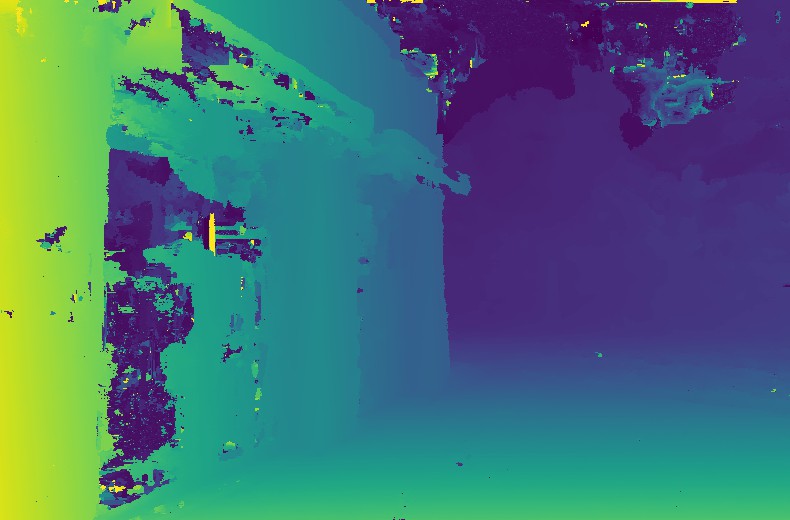}
    \includegraphics[width=0.135\textwidth]{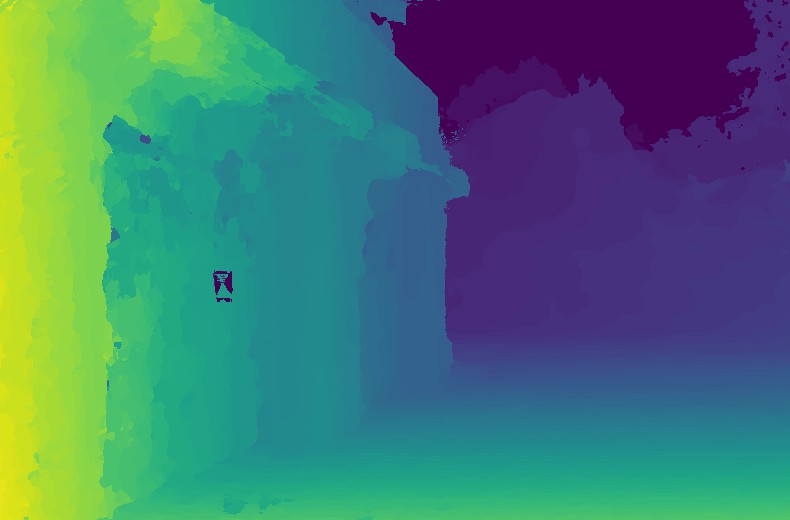} \\
    \vspace{0.5mm}
    \includegraphics[width=0.135\textwidth]{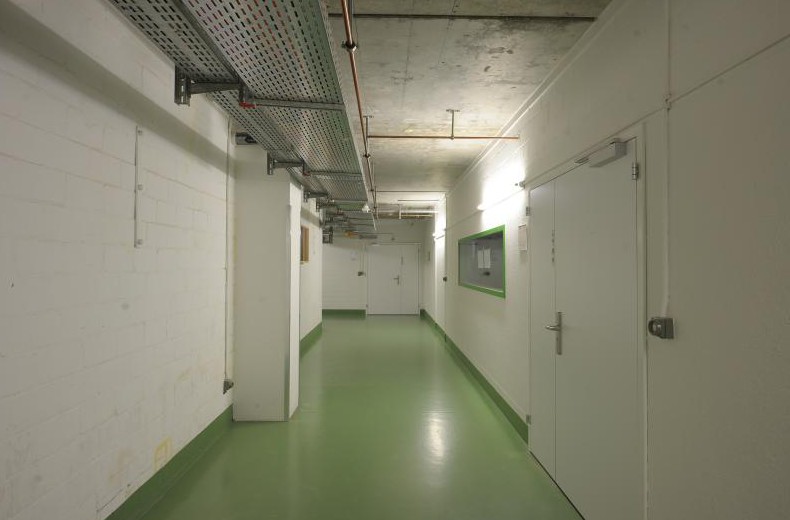}
    \includegraphics[width=0.135\textwidth]{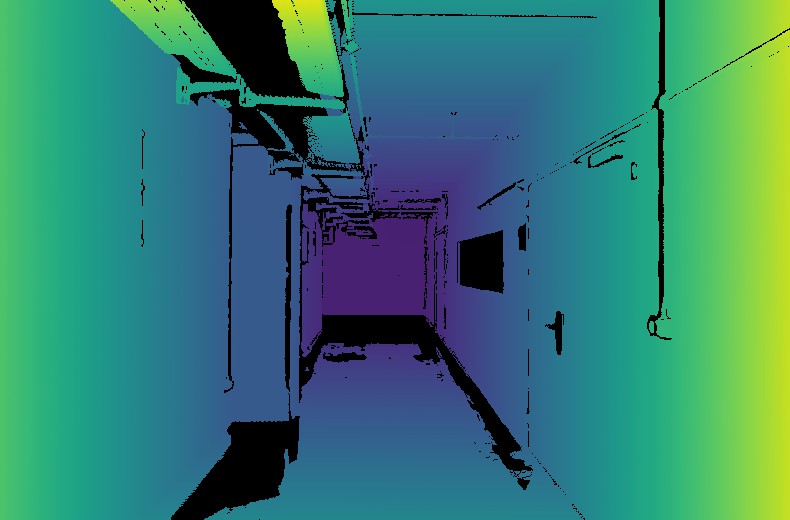}
    \includegraphics[width=0.135\textwidth]{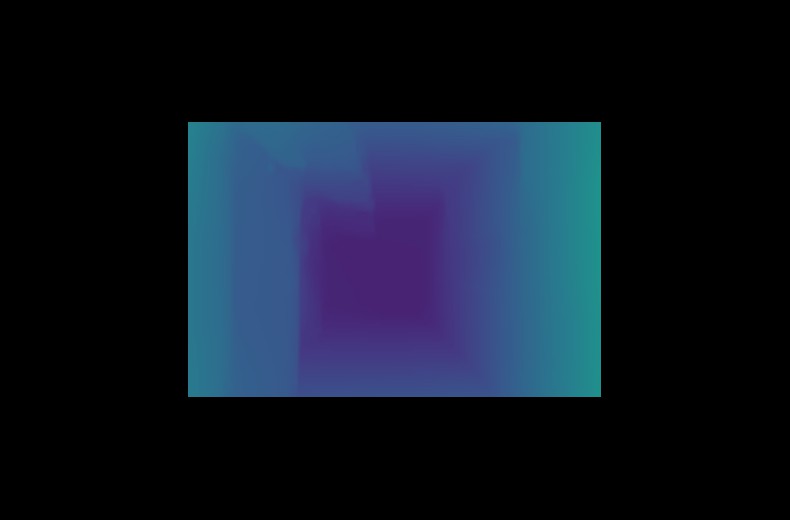}
    \includegraphics[width=0.135\textwidth]{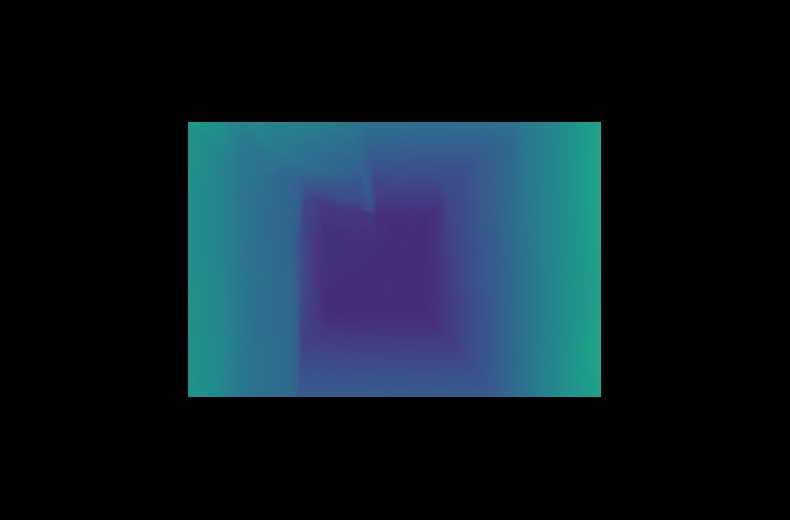}
    \includegraphics[width=0.135\textwidth]{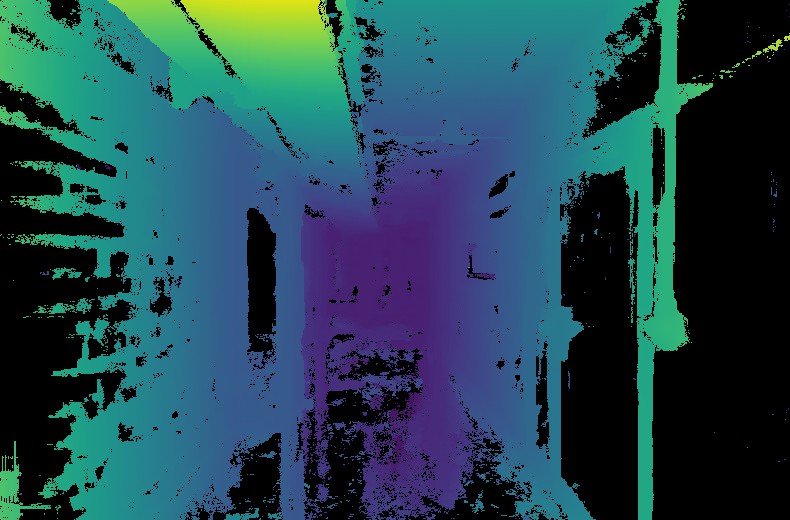}
    \includegraphics[width=0.135\textwidth]{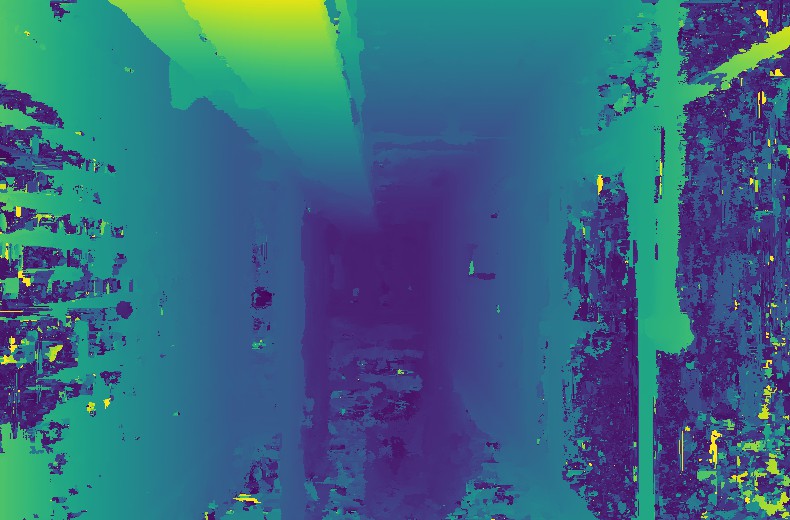}
    \includegraphics[width=0.135\textwidth]{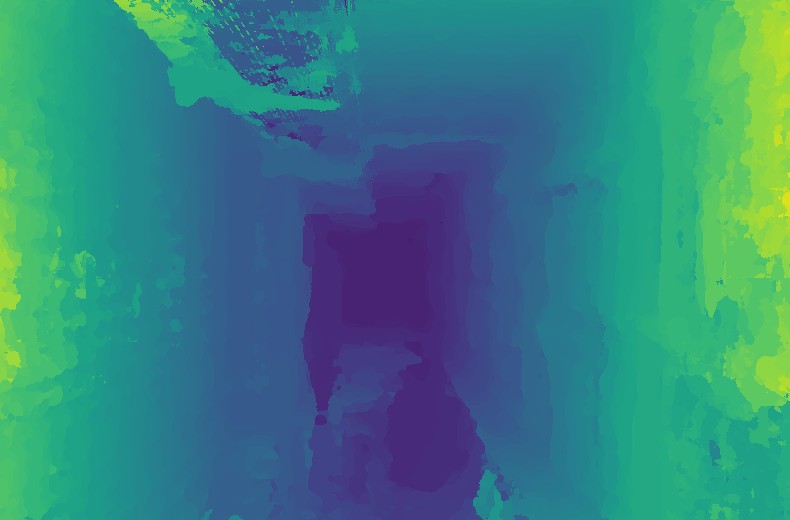}\vspace{0.5mm} \\

\small{
\hfill
\mpage{0.100}{Reference\\\vphantom{(}} \hfill
\mpage{0.100}{Ground truth\\\vphantom{(}} \hfill
\mpage{0.100}{DeMoN \cite{DEMON}\\(best)} \hfill
\mpage{0.100}{DeMoN \cite{DEMON}\\(median)} \hfill
\mpage{0.100}{COLMAP~\cite{COLMAP-MVS} (filtered)} \hfill
\mpage{0.100}{COLMAP~\cite{COLMAP-MVS} (unfiltered)} \hfill
\mpage{0.100}{Our result\\\vphantom{(}}\hfill
}
\vspace{\figcapmargin}
\caption{Qualitative comparisons between different algorithms on ETH3D dataset.}
	\label{fig:comparison}
\end{figure*}

\begin{figure}[t]
	\centering
    \includegraphics[width=0.155\textwidth]{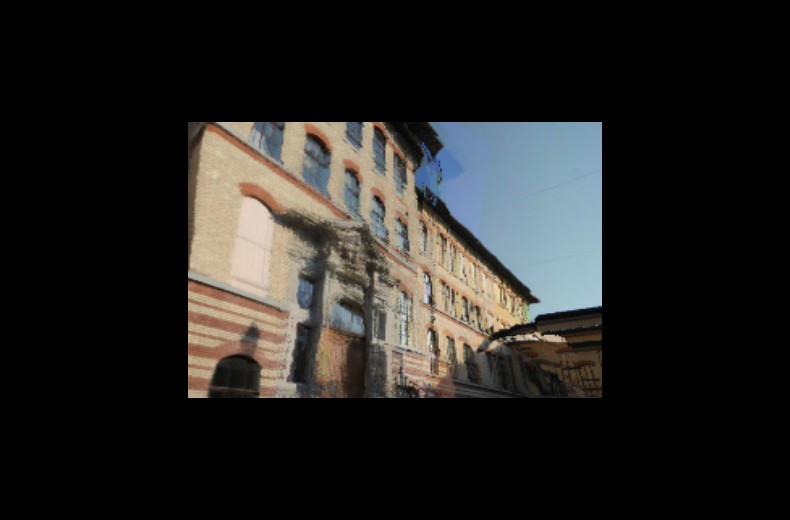}
    \includegraphics[width=0.155\textwidth]{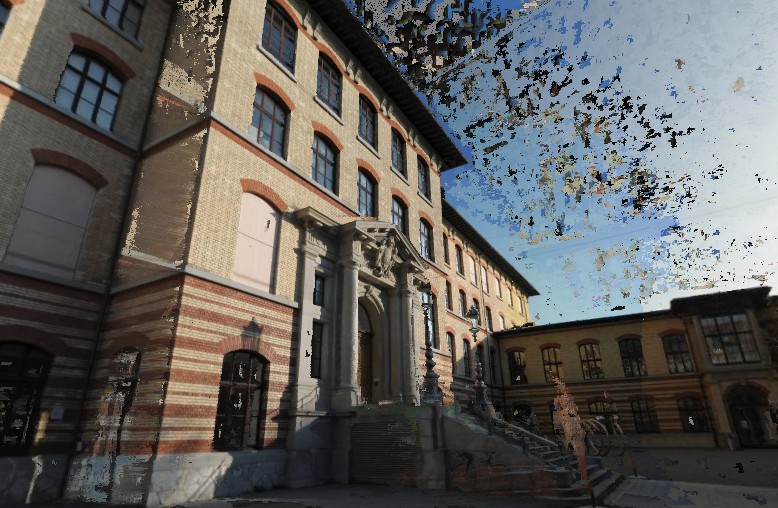} 
    \includegraphics[width=0.155\textwidth]{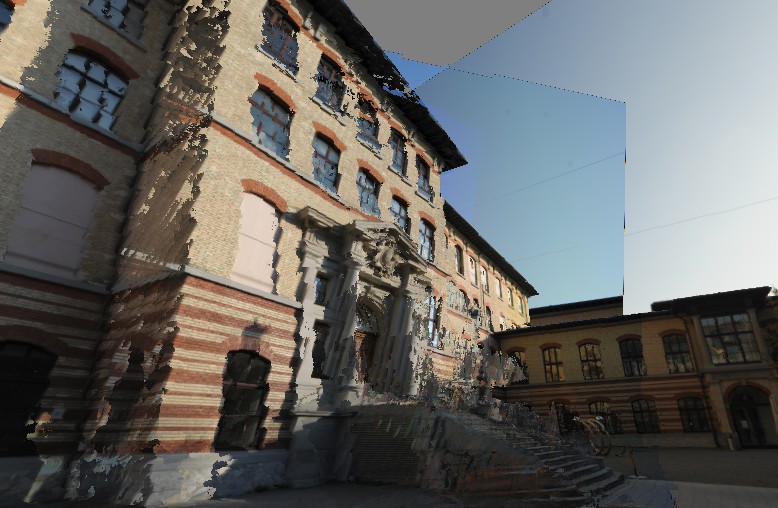}
    \\
    \vspace{0.5mm}
    \includegraphics[width=0.155\textwidth]{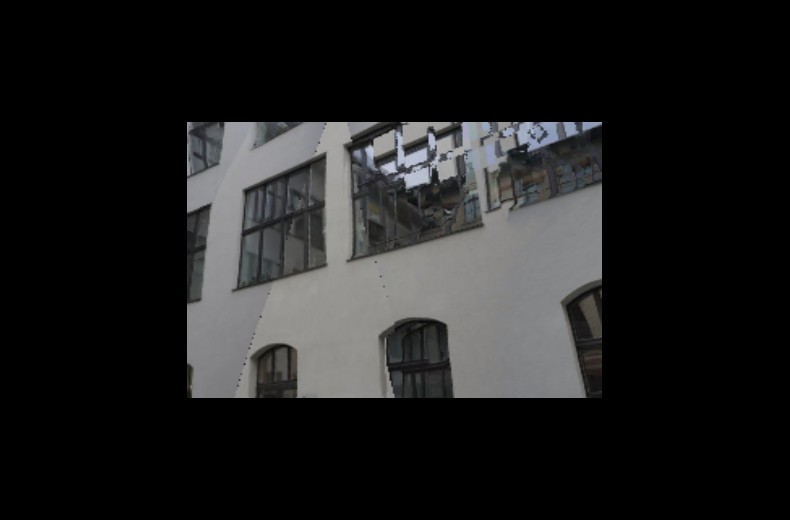} 
    \includegraphics[width=0.155\textwidth]{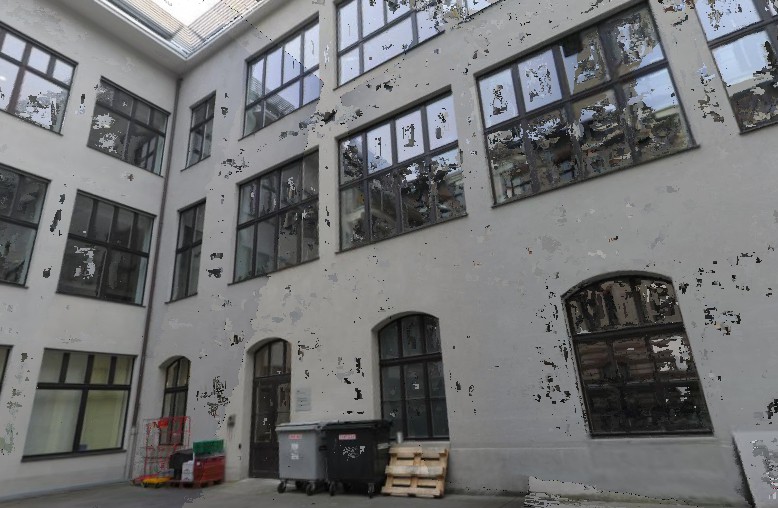} 
    \includegraphics[width=0.155\textwidth]{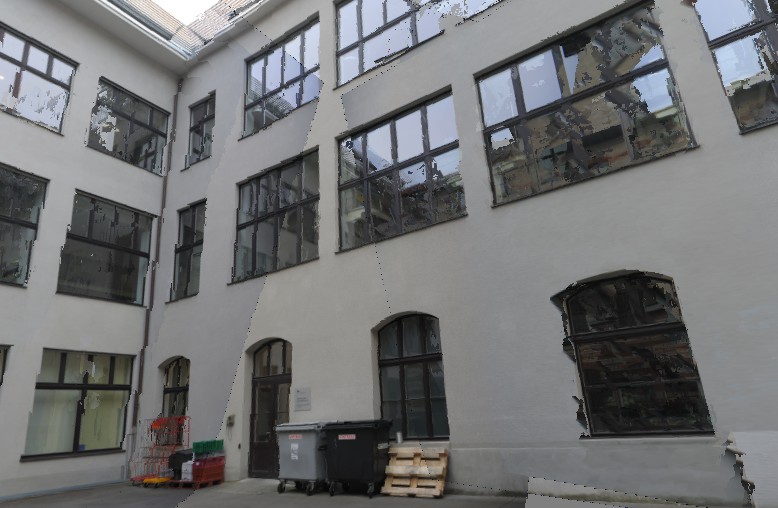} \\
    \vspace{0.5mm}
    \includegraphics[width=0.155\textwidth]{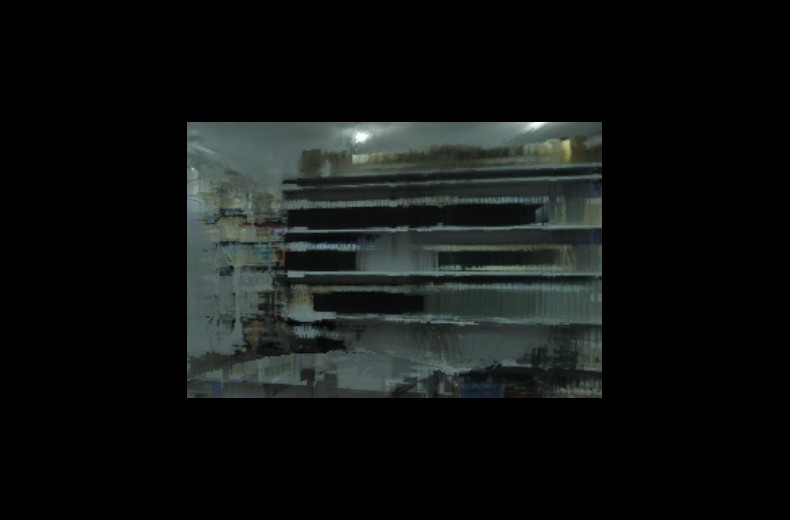}
    \includegraphics[width=0.155\textwidth]{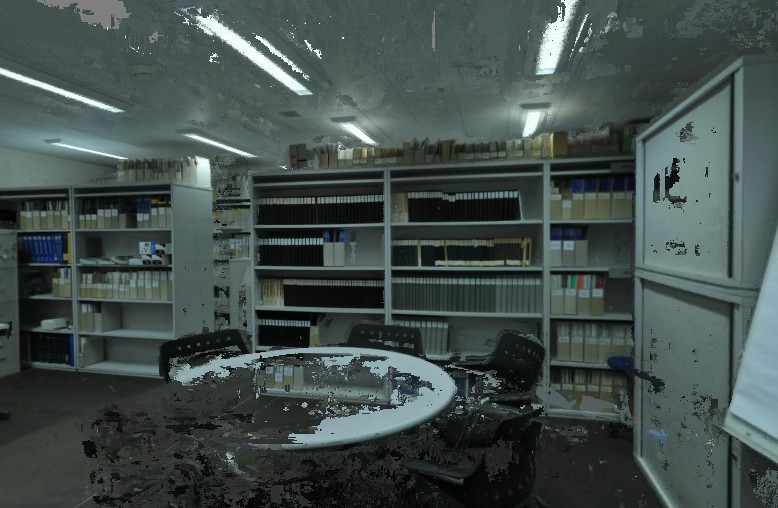}
    \includegraphics[width=0.155\textwidth]{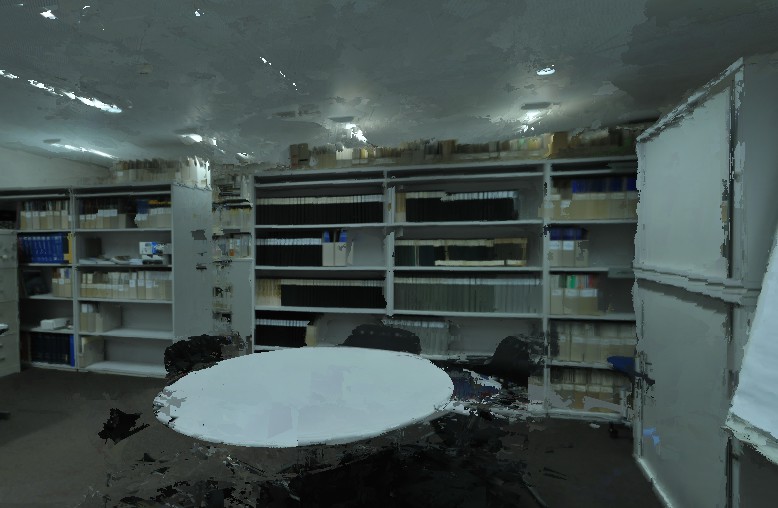} \\
    \vspace{0.5mm}
    \includegraphics[width=0.155\textwidth]{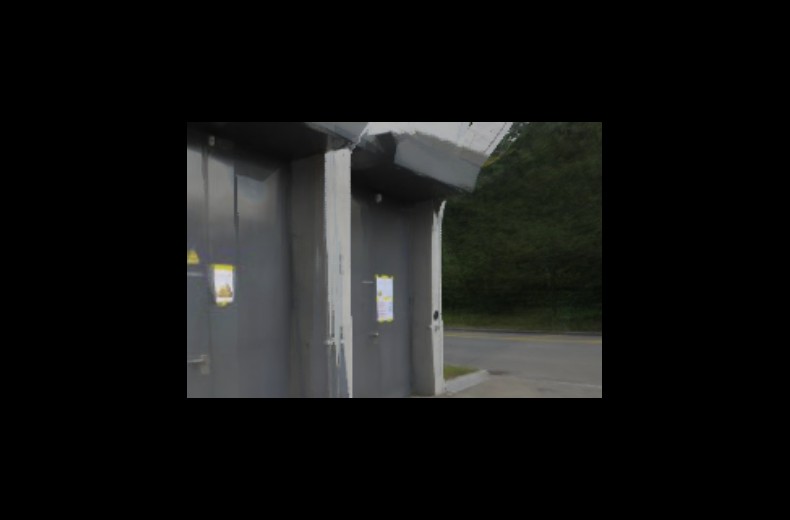}
    \includegraphics[width=0.155\textwidth]{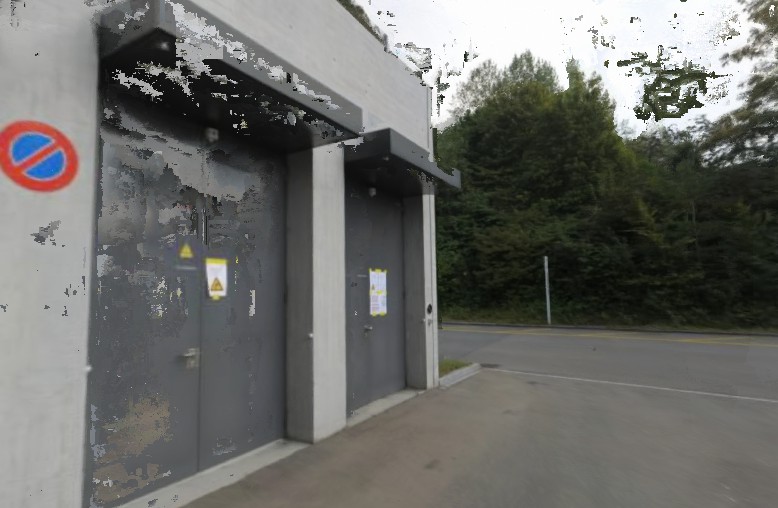}
    \includegraphics[width=0.155\textwidth]{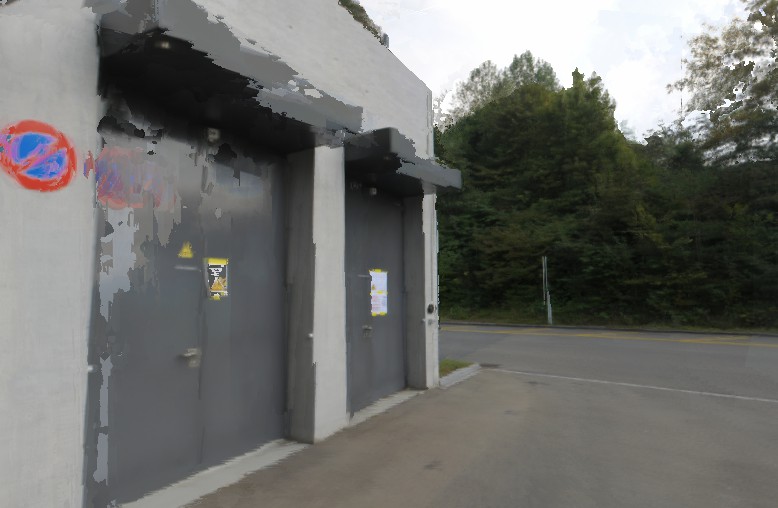} \\
    \vspace{0.5mm}

\mpage{0.31}{DeMoN~\cite{DEMON}\\(best)}\hfill  
\mpage{0.31}{COLMAP~\cite{COLMAP-MVS}\\(unfiltered)}\hfill  
\mpage{0.31}{Our result\\\vphantom{(}}

    \vspace{\figcapmargin}
	\caption{Comparisons of rephotography results. See~\figref{comparison} for the ground truth reference images. 
}
\vspace{\figcapmargin}
\label{fig:comparison-rephoto}
\end{figure}

Since DeMoN is trained with images taken with fixed focal lengths and image resolutions, we crop and resize the images from ETH3D dataset before using them to evaluate DeMoN's performance. This leads to the incomplete reconstruction results in \figref{comparison} and \figref{comparison-rephoto}. The cropped regions are ignored when the error is computed. In addition, DeMoN assumes that the translation between the input image pair is a unit vector. Therefore, we multiply the depth maps produced by DeMoN by the actual translational distance between the two views before comparing them with the ground truths.

\Paragraph{Qualitative comparisons.}
\figref{comparison} shows qualitative comparisons between DeMoN, COLMAP, and our approach. While DeMoN detects the overall structure of the scene, it fails to predict accurate scaling factors and thus results in inaccurate predictions. On the other hand, COLMAP and our approach give accurate predictions wherever the depth cues are sufficient. However, for textureless regions like the sky, the wall, and the surface of the white desks, the predictions made by COLMAP are very noisy, whereas our network is capable of assigning zero disparity to the sky, and interpolating or extrapolating disparities for poorly textured regions.

\figref{comparison-rephoto} shows several rephotography results. The results from DeMoN are often blurry and distorted, indicating that the predictions are not accurate. COLMAP performs well in rephotography in the regions where the predictions are clean. However, for challenging regions, the results contain large holes. Our rephotography results  generally recover the reference images with only small holes. However, edges appear to be jagged because of the disparity quantization  in our approach.

\begin{figure}[t]
	\centering
\mpage{0.48}{\includegraphics[width=\linewidth]{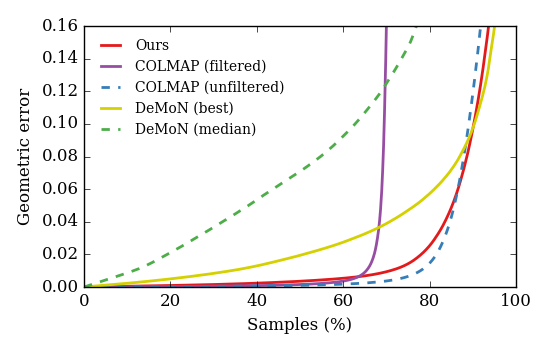}}
\hfill
\mpage{0.48}{\includegraphics[width=\linewidth]{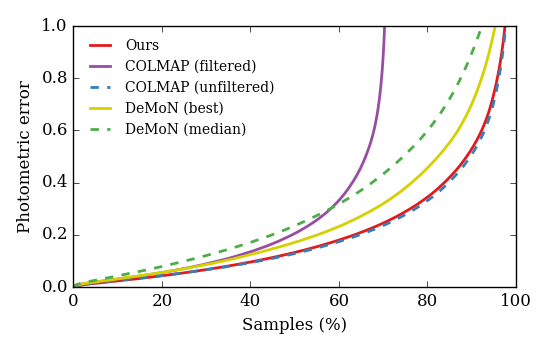}}

\mpage{0.48}{Geometric errors} 
\hfill
\mpage{0.48}{Photometric errors}

\vspace{\figcapmargin}
\caption{The distributions of the errors of different approaches on ETH3D dataset.
}
	\label{fig:comparison-plot}
\end{figure}
\begin{table}[t]
\centering
\caption{Quantitative comparisons between different algorithms on ETH3D dataset.}
\label{tab:comparison}
\vspace{1.0mm}
\resizebox{\linewidth}{!}{%
\begin{tabular}{l|ccc}
\toprule
Algorithm & Completeness & Geo. error & Pho. error \\
\midrule
DeMoN (best) & 100\% & $0.045$ & $0.288$ \\
DeMoN (median) & 100\% &$0.201$ & $0.367$ \\
COLMAP (filtered) & 71\% & $0.007$ & $0.178$ \\
COLMAP (unfiltered) & 100\% & $0.046$ & $\mathbf{0.218}$ \\
Ours & 100\% & $\mathbf{0.036}$ & $0.224$ \\
\bottomrule 
\end{tabular}
}
\end{table}
\Paragraph{Quantitative comparisons.}
\tabref{comparison} shows quantitative comparisons of the average errors over the entire ETH3D dataset between DeMoN, COLMAP, and our approach. First, DeMoN gives much larger errors than COLMAP and our approach with respect to both metrics. COLMAP's filtered predictions have significantly lower average errors, but it discards 29\% of the pixels to achieve that. Finally, COLMAP's unfiltered maps and our results have similar errors. While COLMAP gives slightly lower photometric errors, our approach gives slightly lower geometric errors. 

\figref{comparison-plot} shows the distributions of the errors. We observe that COLMAP predicts 85\% of the pixels with smaller geometric errors than our approach, whereas our approach gives more accurate results for the other 15\% pixels. A possible reason is that for regions with sufficient depths cues, COLMAP produces accurate predictions. Our approach, on the other hand, suffers from the quantized disparity effects. However, for the challenging regions, COLMAP gives noisy predictions which lead to large errors, whereas our approach produces plausible predictions. As for the distributions of the photometric errors, our approach produces almost the same curve as COLMAP does. 

\begin{figure}[t]
	\centering
    \begin{subfigure}[c]{0.28\textwidth}
    	\centering
        \includegraphics[width=0.29\textwidth]{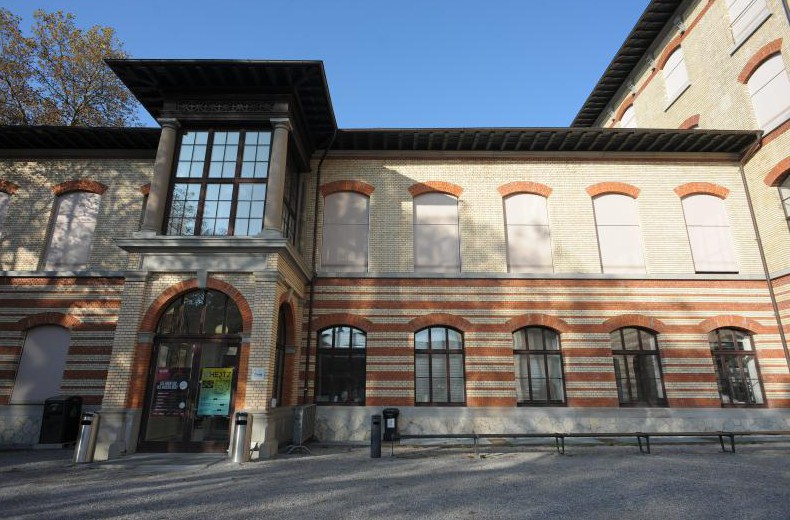}	\hfill
\rotatebox{90}{~\small{N = 1}}
\includegraphics[width=0.29\textwidth]{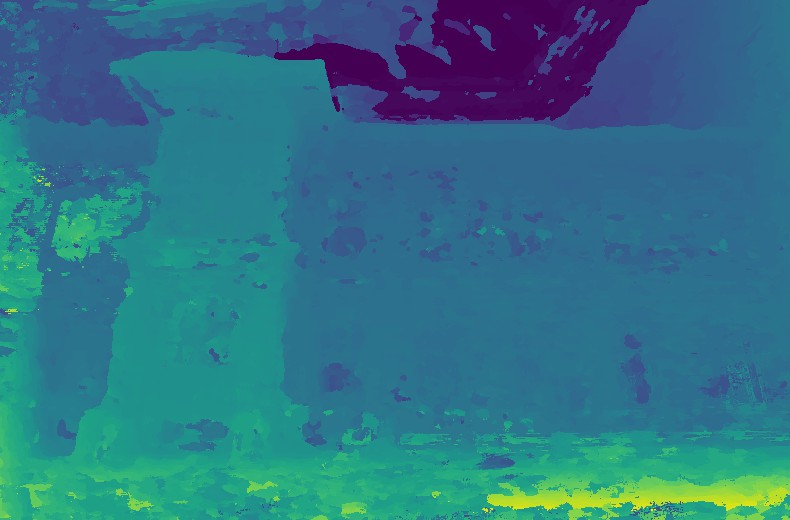}	\hfill
        \includegraphics[width=0.29\textwidth]{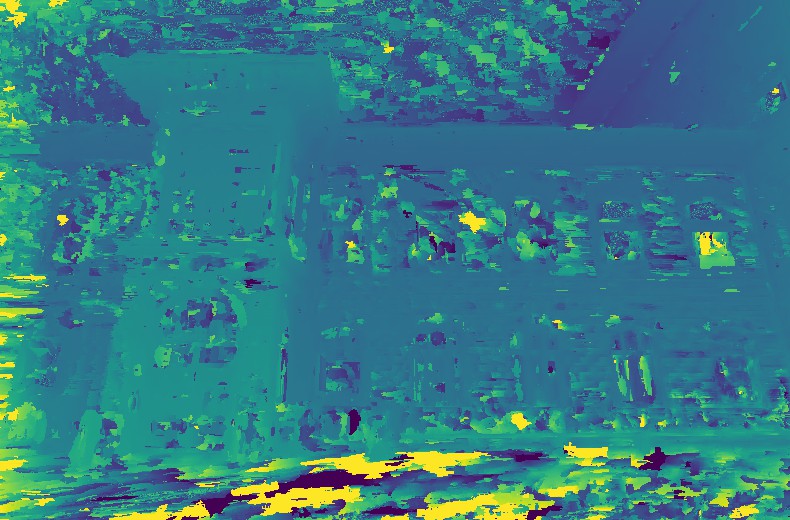}\\
        
        \vspace{0.5mm}
        
        \includegraphics[width=0.29\textwidth]{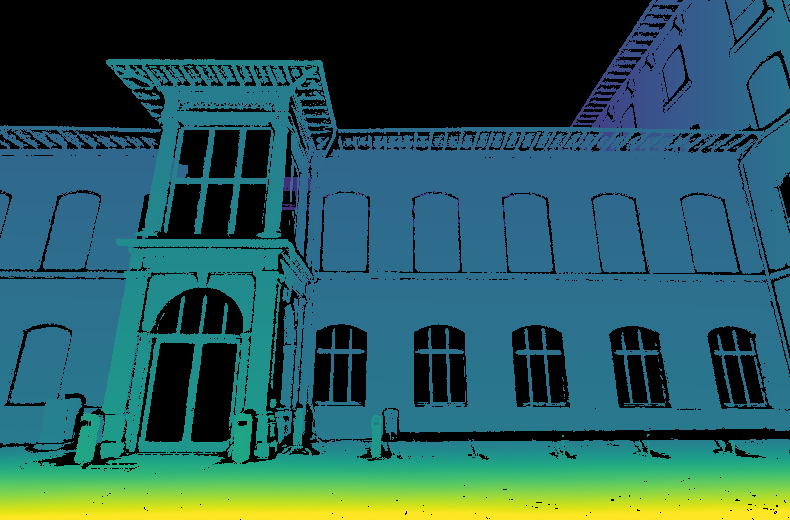} 	\hfill
\rotatebox{90}{~\small{N = 8}}
\includegraphics[width=0.29\textwidth]{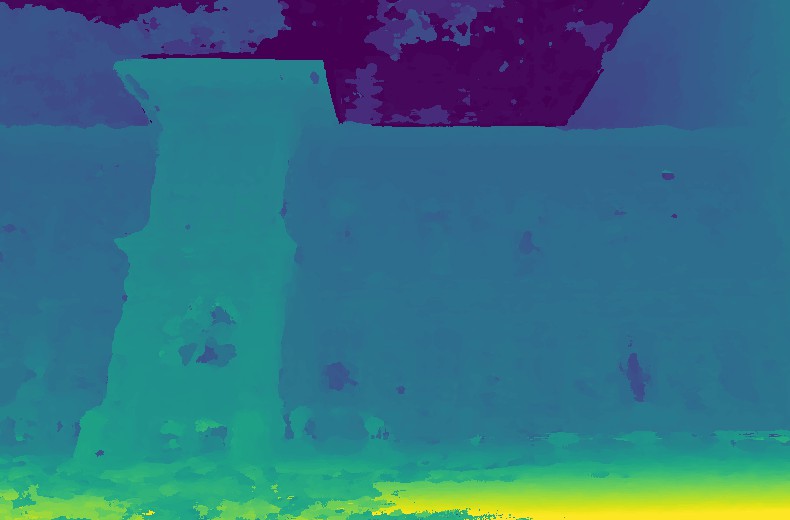}	\hfill
        \includegraphics[width=0.29\textwidth]{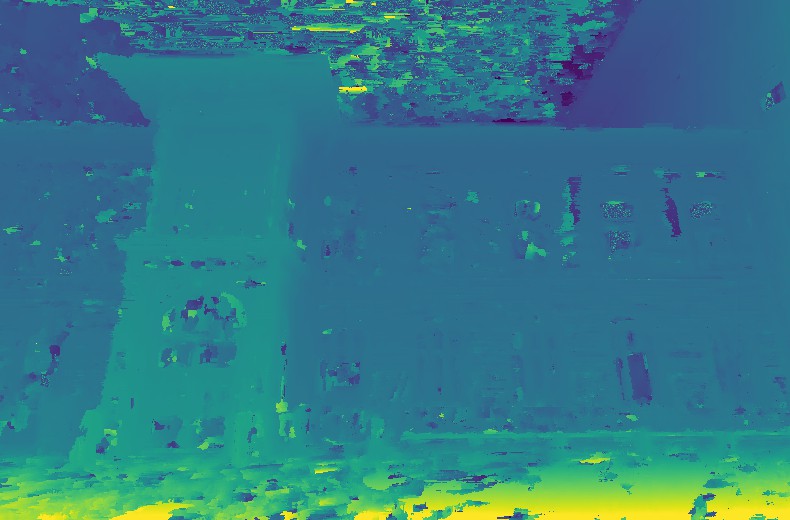}\\

		\vspace{0.5mm}
        
        \includegraphics[width=0.29\textwidth]{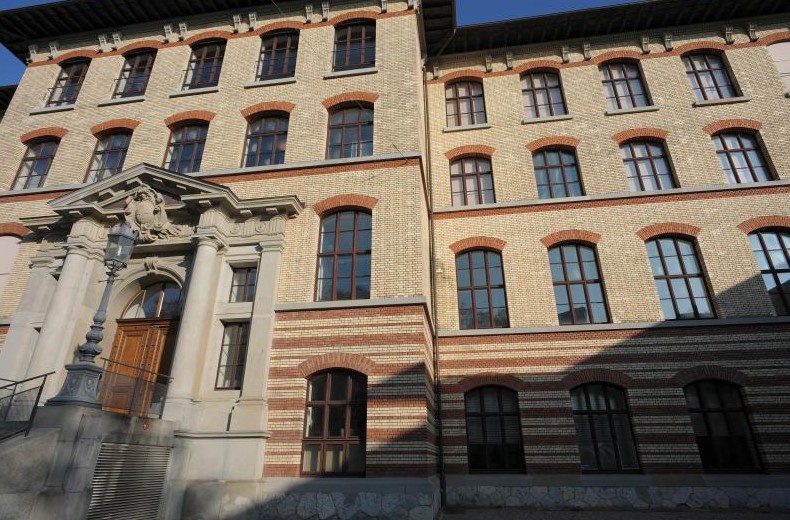} \hfill
\rotatebox{90}{~\small{N = 1}}
        \includegraphics[width=0.29\textwidth]{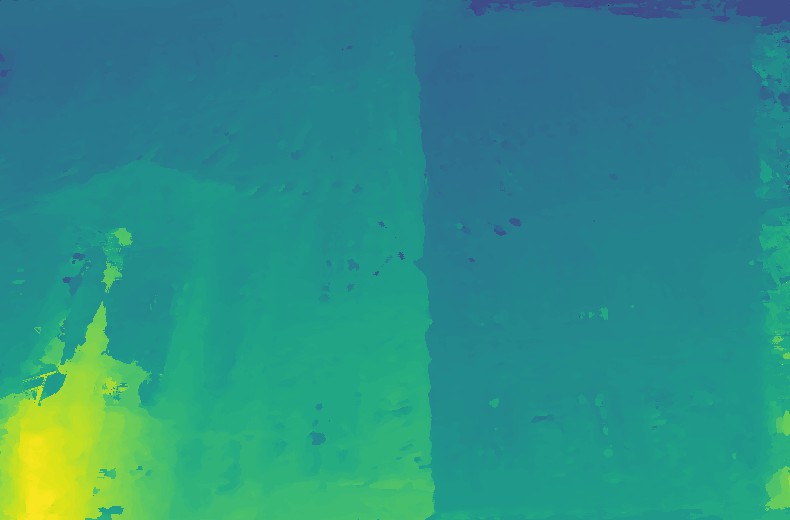} \hfill
        \includegraphics[width=0.29\textwidth]{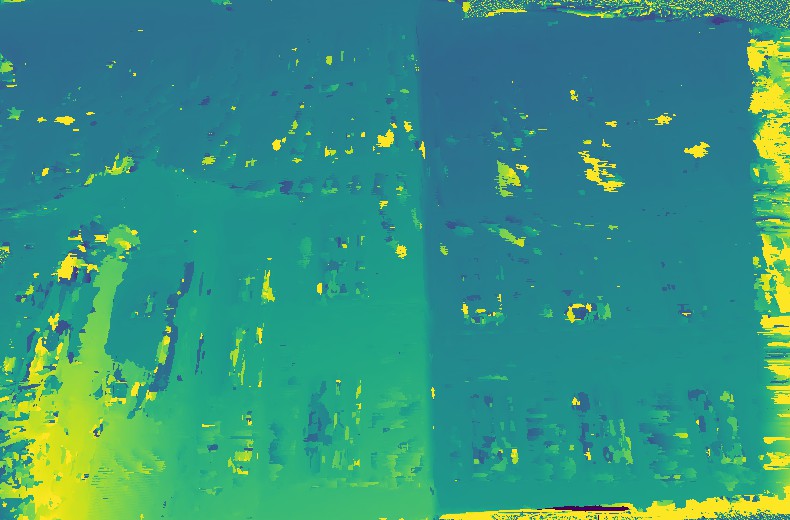} \\
        
        \vspace{0.5mm}
        \includegraphics[width=0.29\textwidth]{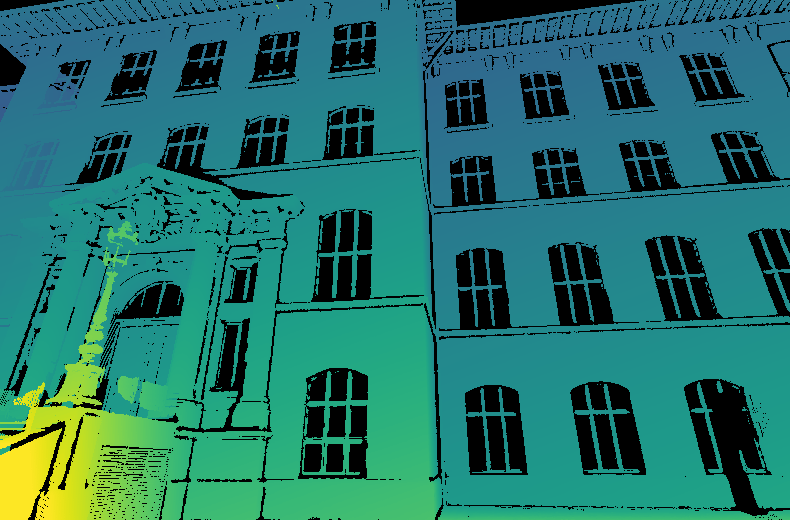} \hfill
        \rotatebox{90}{~\small{N = 8}}
        \includegraphics[width=0.29\textwidth]{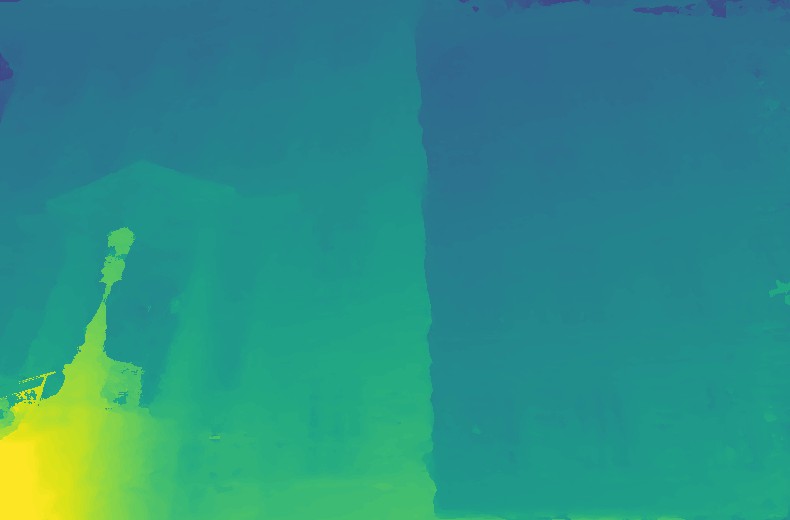}	\hfill
        \includegraphics[width=0.29\textwidth]{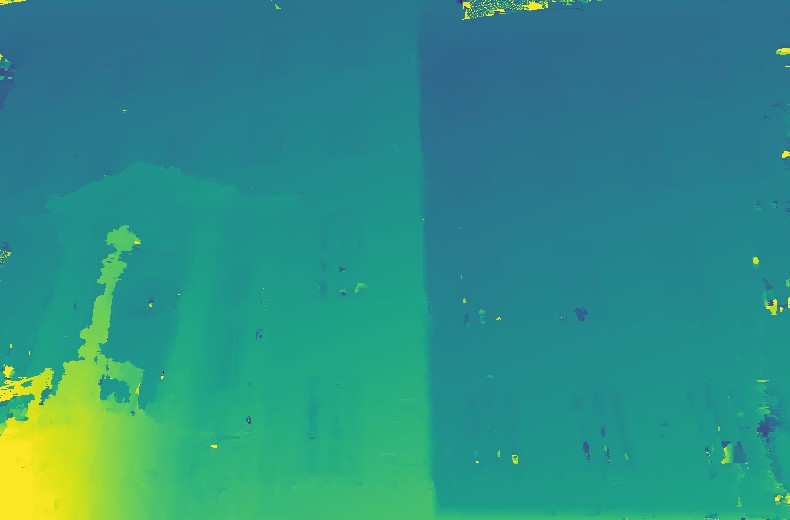} \\

\mpage{0.32}{\footnotesize{Image /\\ground truth}} \hfill
\mpage{0.30}{\footnotesize{Our result\\\vphantom{(}}} \hfill
\mpage{0.26}{\footnotesize{COLMAP~\cite{COLMAP-MVS}\\(unfiltered)}}

    \end{subfigure}
	\begin{subfigure}[c]{0.18\textwidth}
    	\centering
        \includegraphics[width=\textwidth]{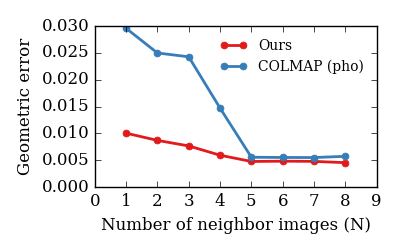}
        \includegraphics[width=\textwidth]{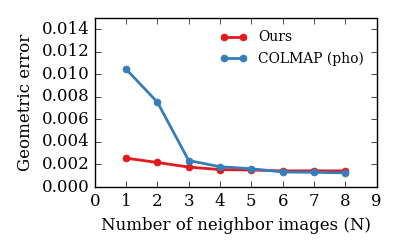} \\
        \footnotesize{Geometric Errors\\\vphantom{(}}
    \end{subfigure}
\vspace{\figcapmargin}
\caption{Examples of progressive improvements for increasing number of input images. 
}
	\label{fig:progressive}
\end{figure}
\Paragraph{Progressive improvement.}
\figref{progressive} shows two examples of the progressive improvements by COLMAP and our approach for an increasing number of input images. When $N$ is small, COLMAP tends to produce large geometric errors, whereas our network can still generate accurate predictions and hallucinate disparities for the regions lacking of good depth cues.

\subsection{Ablation Studies}\label{sec:ablation}

\begin{table}[t]
\centering
\caption{Contributions of different components in our algorithm.}
\label{tab:ablation}
\resizebox{\linewidth}{!}{%
\begin{tabular}{l|cc}
\toprule
Components & Geo. error & Pho. error \\
\midrule
Pretraining & $0.051$ & $0.242$ \\
$+$ U-net & $0.043$ & $0.230$ \\
$+$ U-net $+$ VGG & $0.040$ & $0.226$ \\
$+$ U-net $+$ VGG $+$ DenseCRF & $\mathbf{0.036}$ & $\mathbf{0.224}$ \\
$+$ U-net $+$ VGG $+$ DenseCRF $-$ \textsc{MVS-Synth} & $0.037$ & $0.225$ \\
\bottomrule 
\end{tabular}
}
\end{table}

\Paragraph{DenseCRF.}
\begin{figure}[t]
\centering
\mpage{0.31}{\includegraphics[width=\linewidth]{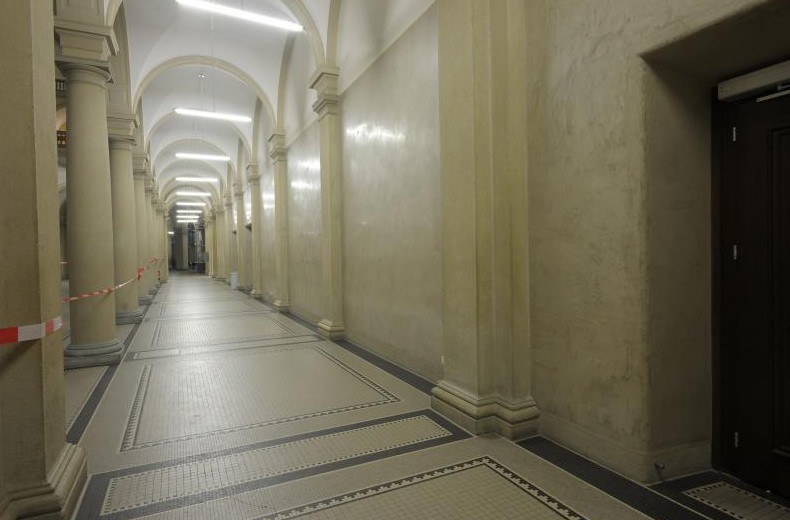}} \hfill
\mpage{0.31}{\includegraphics[width=\linewidth]{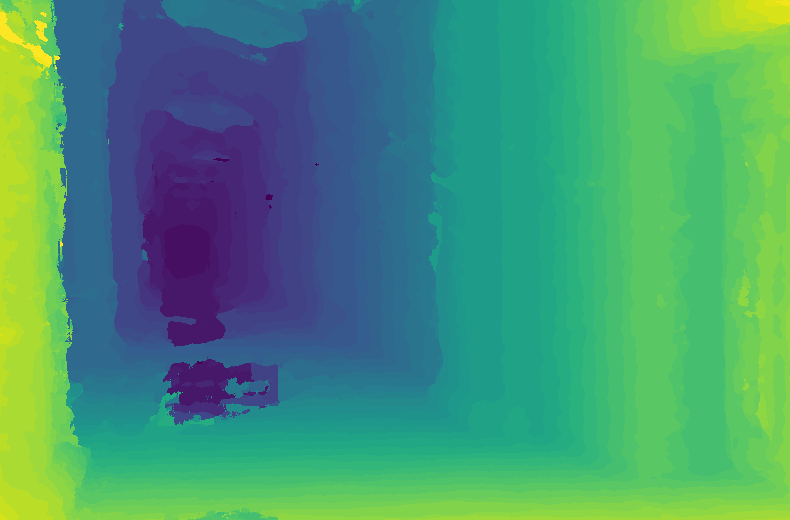}} \hfill
\mpage{0.31}{\includegraphics[width=\linewidth]{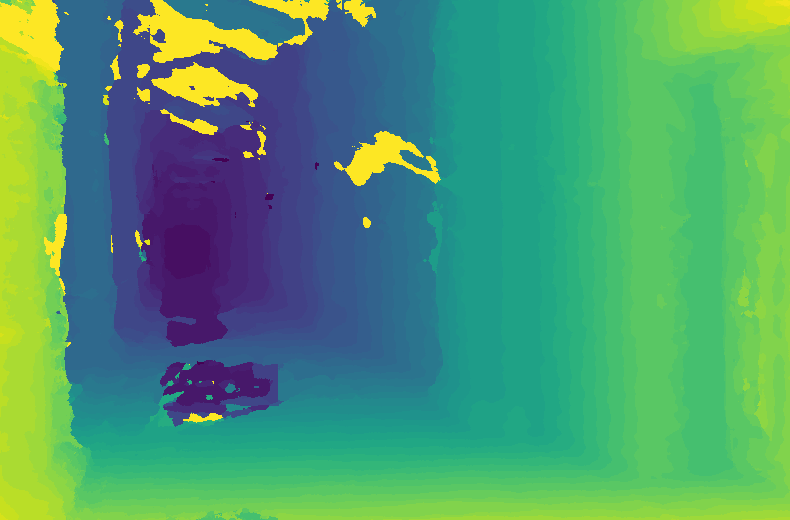}}

\vspace{1mm}

\mpage{0.31}{Image} \hfill
\mpage{0.31}{Ours} \hfill
\mpage{0.31}{Ours w/o DenseCRF} 

\vspace{\figcapmargin}
\caption{Example of the improvements from the DenseCRF refinement. Applying DenseCRF removes the noisy predictions.}
\label{fig:ablation-CRF}
\end{figure}
As shown in \figref{ablation-CRF}, applying DenseCRF removes a large portion of the noisy patches in low-confidence regions such as the reflective wall, and encourages the disparity predictions to follow the color edges. As shown in~\tabref{ablation}, DenseCRF improves the results with respect to both error metrics.

\begin{figure}[t]
	\centering
\mpage{0.31}{\includegraphics[width=\linewidth]{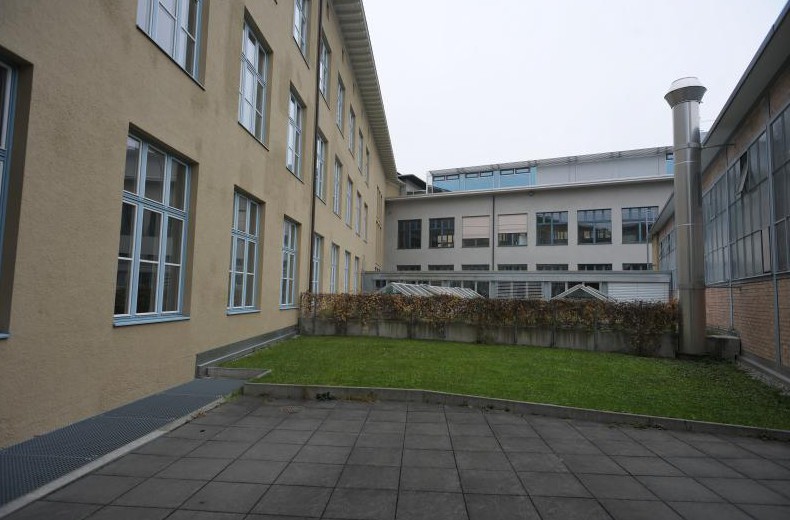}} \hfill
\mpage{0.31}{\includegraphics[width=\linewidth]{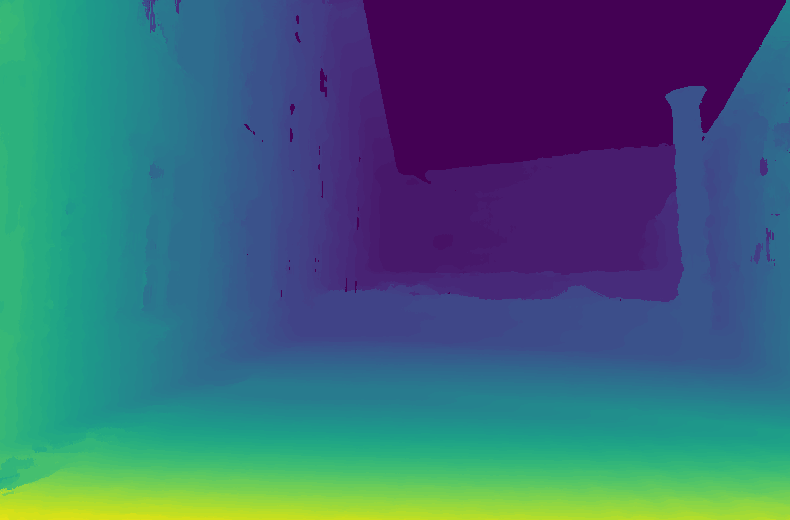}} \hfill
\mpage{0.31}{\includegraphics[width=\linewidth]{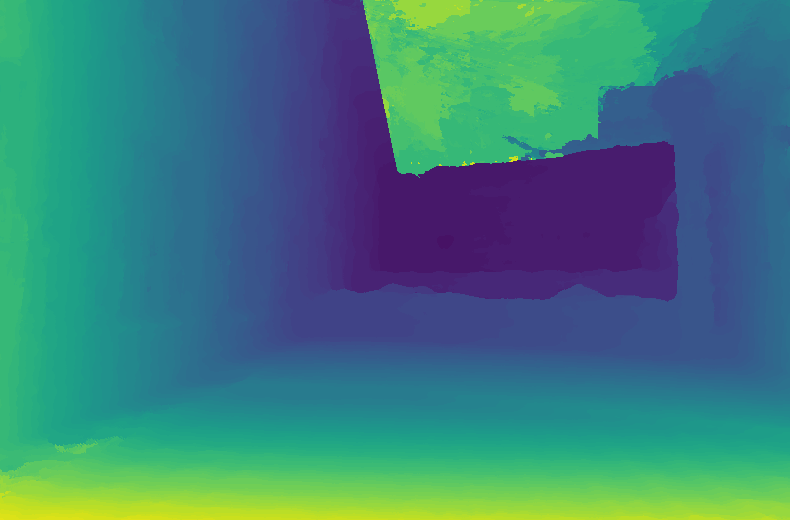}}

\vspace{1mm}

\mpage{0.31}{Image} \hfill
\mpage{0.31}{Ours} \hfill
\mpage{0.31}{Ours w/o \textsc{MVS-Synth}} 

\vspace{\figcapmargin}
\caption{Comparisons between networks trained with and without the \textsc{MVS-Synth} dataset. Without \textsc{MVS-Synth} dataset, the network has difficulty in handling regions such as the sky because real-world datasets do not cover these regions.
%
}
	\label{fig:ablation-GTAV}
\end{figure}
\Paragraph{MVS-Synth dataset.}
\tabref{ablation} shows that removing \textsc{MVS-Synth} dataset from the training set results in slightly larger errors for both metrics. Qualitatively, we observe that the network trained without \textsc{MVS-Synth} dataset works very poorly for the sky and reflective surfaces, as~\figref{ablation-GTAV} shows. These regions usually lack ground truth data, so the errors do not reflect much on the quantitative errors. We suggest that the poor predictions result from the fact that the ground truths in DeMoN dataset does not cover such regions.


\Paragraph{U-Net and VGG features.}
As~\tabref{ablation} shows, adding the U-net and VGG features each provides improvements in both error metrics. This shows that allowing non-local information and providing semantic features both help the network in better disparity predictions.

\subsection{Limitations}
Following are some limitations of our network. First, the quantization of disparity results in undesired geometric and photometric errors. Second, our network often fails to predict correct disparities for vegetation areas containing trees or grass. 
Finally, the computation speed of our algorithm is constrained by the time-consuming generation of the plane-sweep volumes and the deep and large network structures.
\section{Conclusions}

With DeepMVS, we demonstrate the feasibility of learning Mulit-View Stereopsis with a convolutinoal neural network, and show that learning-based approaches can overcome the weaknesses of conventional algorithms. 
\section*{Acknowledgement}
\vspace{-2mm}
We gratefully acknowledge the support of NVIDIA Corporation with the donation of the GPUs used in this research and the grant from Office of Naval Research N00014-16-1-2314.

{\small
\bibliographystyle{ieee}
\bibliography{mvsbib}
}

\end{document}